\documentclass[10pt,twocolumn,letterpaper]{article}

\usepackage[pagenumbers]{wacv}              % To produce the CAMERA-READY version
\newcommand{\methodname}{Instructive3D}

\usepackage{graphicx}
\usepackage{amsmath}
\usepackage{amssymb}
\usepackage{booktabs}
\usepackage[pagebackref,breaklinks,colorlinks]{hyperref}

\usepackage[capitalize]{cleveref}
\crefname{section}{Sec.}{Secs.}
\Crefname{section}{Section}{Sections}
\Crefname{table}{Table}{Tables}
\crefname{table}{Tab.}{Tabs.}

\def\wacvPaperID{2230} % *** Enter the WACV Paper ID here
\def\confName{WACV}
\def\confYear{2025}

\begin{document}

\title{Instructive3D: Editing Large Reconstruction Models with Text Instructions}

\author{
	Kunal Kathare\thanks{Equal Contribution} \quad
	Ankit Dhiman$^{*}$ \quad
	K Vikas Gowda \quad 
	Siddharth Aravindan \quad
	Shubham Monga \quad \\
	Basavaraja Shanthappa Vandrotti \quad
	Lokesh R Boregowda \\
	Samsung R\&D Institute India - Bangalore
}
\maketitle

%%%%%%%%% ABSTRACT
\begin{abstract}
   % In the domain of 3D reconstruction, currently existing large reconstruction models perform exceptionally well in 
   % generating high quality 3d objects given a single input image. However, these models lack mechanisms for fine grained control over 
   % the generated 3d objects. This paper presents InstructTri2Tri, a novel approach designed to integrate fine grained control into 
   % existing Large reconstruction models, by adding conditioning using text prompts in the triplane latent space.
   % The objective of this research is to enhance the versatility and precision of large reconstruction models by allowing users to 
   % specify modifications through text prompts. Our approach leverages the robust capabilities of the existing large reconstruction
   % models while introducing a text based editing layer that applies user defined instructions. Experimental results demonstrate 
   % that InstructTri2Tri effectively enables nuanced adjustments to 3D objects, thus significantly expanding the creative and 
   % practical applications of automated 3D reconstruction technology. The implications of this advancements suggest a broader 
   % scope for customization and user interaction in 3D model generation, paving the way for more intuitive and precise design workflows.
	
	Transformer based methods have enabled users to create, modify, and comprehend text and image data. Recently proposed Large Reconstruction Models (LRMs) further extend this by providing the ability to generate high-quality 3D models with the help of a single object image. These models, however, lack the ability to manipulate or edit the finer details, such as adding standard design patterns or changing the color and reflectance of the generated objects, thus lacking fine-grained control that may be very helpful in domains such as augmented reality, animation and gaming. Naively training LRMs for this purpose would require generating precisely edited images and 3D object pairs, which is computationally expensive. In this paper, we propose~\methodname{}, a novel LRM based model that integrates generation and fine-grained editing, through user text prompts, of 3D objects into a single model. We accomplish this by adding an adapter that performs a diffusion process conditioned on a text prompt specifying edits in the triplane latent space representation of 3D object models. Our method does not require the generation of edited 3D objects. Additionally,~\methodname{} allows us to perform geometrically consistent modifications, as the edits done through user-defined text prompts are applied to the triplane latent representation thus enhancing the versatility and precision of 3D objects generated. We compare the objects generated by~\methodname{} and a baseline that first generates the 3D object meshes using a standard LRM model and then edits these 3D objects using text prompts when images are provided from the Objaverse LVIS dataset. We find that~\methodname{} produces qualitatively superior 3D objects with the properties specified by the edit prompts.
\vspace{-7mm}
\end{abstract}

\begin{figure}[!t]
	\centering
	\includegraphics[width=\linewidth]{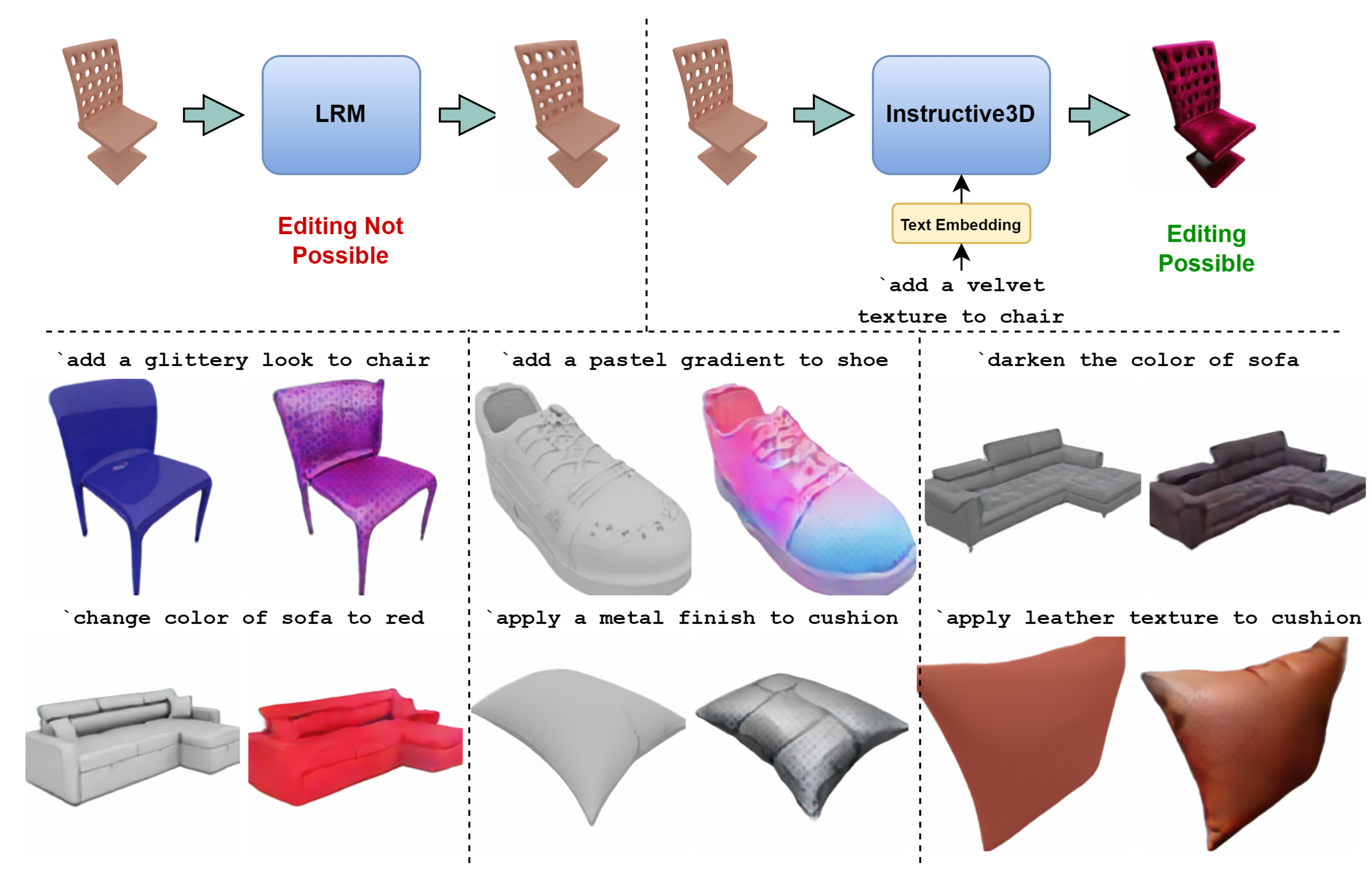}
	\caption{\textbf{An Overview of~\methodname{}.} The top section illustrates the limitations of existing large reconstruction models (LRMs), which lack the capability for fine-grained control over generated 3D objects. In contrast, the bottom section presents examples of how Instructive3D enables fine-grain control to 3D models using text-based prompts, showcasing the enhanced versatility and control offered by our approach.}
	\label{fig:teaser}
	\vspace{-7mm}
\end{figure}

\addtocontents{toc}
{\protect\setcounter{tocdepth}{-2}}
\section{Introduction}
\noindent
The increasing popularity of augmented reality (AR) and virtual reality (VR) applications has resulted in demand for the generation of 3D content. Generation of high-quality 3D content, however, is a nontrivial task that not only requires specific domain knowledge but can be labor intensive. The development of techniques for the generation of 3D content, such as objects, from minimal cues like image(s) of the object, is an emerging field of research. Recent works including~\cite{Kanazawa2018LearningCM, Zhang2022Monocular3O} rely on keypoints as well as 3D Generative Adversarial Network (GAN) priors~\cite{Goodfellow2014GenerativeAN} to generate the 3D object shapes from a single image. Additionally, recent methods, such as~\cite{yu2021pixelnerf, Charatan2023pixelSplat3G}, use Neural Radiance Fields (NeRFs)~\cite{mildenhall2021nerf} and 3D Gaussian splatting~\cite{kerbl20233d} to learn the 3D representations and incorporates the underlying geometric details to generate photorealistic novel views of desired objects. However, these methods rely on multi-view images, which are not always available. 

\noindent
Furthermore, numerous recent approaches~\cite{poole2022dreamfusion, Wang2023ProlificDreamerHA, tang2024dreamgaussian, yi2024gaussiandreamer, xu2024agg, zhou2024gala3dtextto3dcomplexscene} utilize sampling techniques: Score distillation sampling (SDS)~\cite{poole2022dreamfusion} and variational score distillation (VSD)~\cite{Wang2023ProlificDreamerHA} to leverage the pre-trained text-to-image diffusion models for the generation of 3D representations from a text prompt or a single image of the object. However, these methods require test-time optimization, increasing the computational requirements for object generation.

\noindent
Recent advancements in Natural Language Processing~\cite{Brown2020LanguageMA} and 2D Computer Vision~\cite{radford2021learning} have assisted in building foundational models for single-view 3D reconstruction.  Large Reconstruction Model (LRM)~\cite{hong2024lrm}, is one such model that uses a transformer-based encoder-decoder architecture to learn 3D representations of objects from a single image. It does so by first utilizing a pre-trained vision transformer DINO~\cite{Caron2021EmergingPI} to generate image features, which are projected to a 3D triplane via a cross-attention image-to-triplane decoder. These triplane features are then first rendered from an arbitrary view using 3D representation methods such as NeRFs and then decoded to obtain the 3D object shape and texture. Akin to other foundational models, LRMs can reconstruct high-fidelity 3D shapes from a wide range of images captured in the real world with a lower inference time.

\noindent
In addition to the generation of 3D objects, approaches that can edit the style and texture of an object, while preserving the underlying geometry are also useful for several applications including gaming and animation in AR and VR. These methods can help users make detailed adjustments to the shape, texture, or features of the 3D object helping them meet specific design criteria or aesthetic preferences of the user. The absence of such control mechanisms in existing LRM models poses a significant barrier to their boarder adoption in creative and professional fields.

\noindent
In this work, we propose \textbf{\methodname{}}, a novel text-based editing approach that integrates fine-grained object editing control into the existing LRM methods.~\methodname{} achieves this by incorporating an extra layer of interaction where users can specify edits through natural language text prompts. These instructions are then incorporated into the reconstruction process by conditioning the triplane latent features with the text prompt using a latent diffusion process. Additionally, we leverage the abilities of the existing foundational model to generate the training data for~\methodname{}. We use InstructPix2Pix~\cite{brooks2023instructpix2pix} to generate edited object images given an image and manually curated text prompts. Given the edited images, the triplane latent features of the edited objects are then procured using a pre-trained LRM to supervise the latent diffusion process.~\methodname{} enables users to refine generated 3D objects through text prompts, demonstrating the potential of the proposed diffusion-based text prompt conditioned editing. This improves the flexibility and usability of the LRM models. Enabling fine-grained control has the potential to transform the way 3D models are created and manipulated, as it provides new avenues for innovation and creativity in various domains. In summary, our main contributions are as follows:

\begin{itemize}
	\item We introduce \methodname{}, a novel approach for integrating fine-grained control into the reconstruction process of existing LRM architectures. We achieve this through text-conditioning of the triplane features of LRM architectures.
	\item This text-conditioning is done with the help of a latent diffusion process that operates in the triplane latent space of the 3D object.
	\item We also propose a novel data-curation pipeline to generate the dataset to train \methodname{}, as acquiring edited 3D object pairs is very difficult. We leverage the abilities of existing foundational models to do so.
	\item Additionally, we showcase the superior generation quality of 3D objects generated by \methodname{} over other baselines with various editing prompts on standard 3D object datasets such as Objaverse.
\end{itemize}
\begin{figure*}[h]
	\centering
	\includegraphics[width=\linewidth]{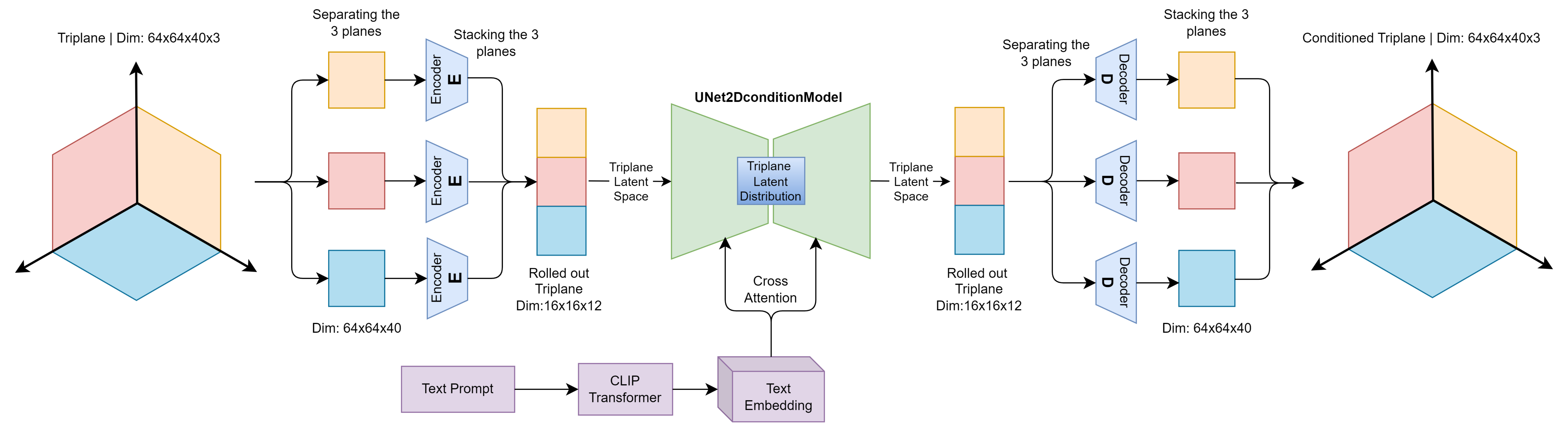}
	\caption{\textbf{The architecture of our adapter ~\methodname{}.} The triplane is first generated by the LRM (in this case Real3D);  each plane of the triplane is then separated, normalized between [-1,1] and processed through its dedicated encoder, trained specifically for the corresponding plane. The resulting latent planes have their channels concatenated and passed through a conditional UNet~\cite{ronneberger2015unetconvolutionalnetworksbiomedical} model for denoising, in conjunction with a text embedding obtained from a CLIP~\cite{radford2021learning} transformer based on the input text prompt. The denoised output is then separated back into the three planes, which are passed through their respective decoders. Finally, the planes are stacked together to form the conditioned triplane, reflecting the user-specified text-based modifications.}
	\label{fig:architecture}
	\vspace{-5mm}
\end{figure*}

%This paper outlines the development and implementation of~\methodname{}, detailing the underlying methodologies and the integration of text-based control. We present comprehensive experimental results to validate the effectiveness of our approach and discuss its implications for the future of 3D model generation and customization.In the following sections, we will delve into the related work in 3D reconstruction and text-based editing, describe the methodology employed in developing~\methodname{}, present our experimental results, and discuss the implications of our findings for future research and applications in the field of 3D generation.

\section{Background and Related Work}
\label{sec:rel_work}

\subsection{Diffusion Models}
\noindent
Diffusion models learn the distribution of an input by iterative denoising. In the forward diffusion process, a Gaussian noise $\epsilon \sim \mathcal{N} \left( 0,1 \right)$  is sequentially added for $T$ timesteps to a clean sample $x_0$ to get a noisy sample  $x_T$.  In the backward process, the clean sample  $x_0$ is retrieved back by iterative denoising of the noisy sample  $x_T$ for the same number of timesteps. The iterative denoising process is modelled with a denoising network $\epsilon_{\theta}$ conditioned on the timestep  $t\in\{1,T\}$ and optional conditioning  $c$ (e.g. text prompts, depth-maps, normal-maps, etc.). The denoiser network is trained with a simple mean-squared loss:
\begin{equation}
	L_{D} = \;\; E_{x_0,\epsilon \sim \mathcal{N} \left( 0,I \right) ,t} ||\epsilon -\epsilon _{\theta}\left(z_t, t, c\right)||^2 
	\label{eq:diff_loss}
\end{equation}

\noindent
\textbf{Utilizing Pre-trained 2D Diffusion Models} 
Methods like DreamFusion~\cite{poole2022dreamfusion} use a pre-trained 2D text-to-image model to perform text-to-3D synthesis using SDS sampling. These methods use NeRF for the 3D generation task which is slow to train. Recently, DreamGaussian~\cite{tang2024dreamgaussian} utilizes fast 3DGS representation instead of NeRF and then converts 3D Gaussians into textured meshes followed by a fine-tuning stage to refine the details. GaussianDreamer~\cite{yi2024gaussiandreamer} utilizes a 3D diffusion model as a prior for initialization and uses a 2D diffusion model to enrich the geometry and the appearance. AGG~\cite{xu2024agg} instantly produces 3d objects from a single image, eliminating the need for per-instance optimization, AGG is a cascaded pipeline that first generates a coarse representation of 3D data and later up-samples it with a 3D Gaussian super-resolution module. GALA3D~\cite{zhou2024gala3dtextto3dcomplexscene} is a layout-guided Gaussian splatting framework for complex text to 3D generation. It bridges text description and compositional scene generation through layout priors obtained from LLMs and a layout refinement module that optimizes the coarse layout interpreted by the LLMs.

\noindent
\textbf{Diffusion Models \& Multimodal 3D.} Previous works have tried to use pre-trained 2D diffusion models for 3D generation and to generate multiple novel views and then use NeRF~\cite{mildenhall2020nerf} for volumetric rendering, but these suffer from issues like the Janus effect and require test time optimization. One such work is Score Jacobian Chaining~\cite{wang2023score}, which applies the chain rule on learned gradients, and back propagates the score of a diffusion model through the Jacobian of a differentiable renderer, which they instantiate to be a voxel randiance field. This framework collects 2D scores at multiple camera viewpoints into a 3D score, reusing a pre-trained 2D model for 3D generation. Another such work is Zero1to3~\cite{liu2023zero1to3} which changes the camera viewpoint of an object given just a single RGB image. It uses geometric priors of large-scale diffusion models, to change the camera viewpoint when given a single RGB image by learning it on a synthetic dataset, then uses NeRF for volumetric rendering, this model uses Score Distillation Sampling which suffers from slow per instance optimization, which in turn increases the inference time.

\begin{figure*}[!t]
	\centering
	\includegraphics[width=\linewidth]{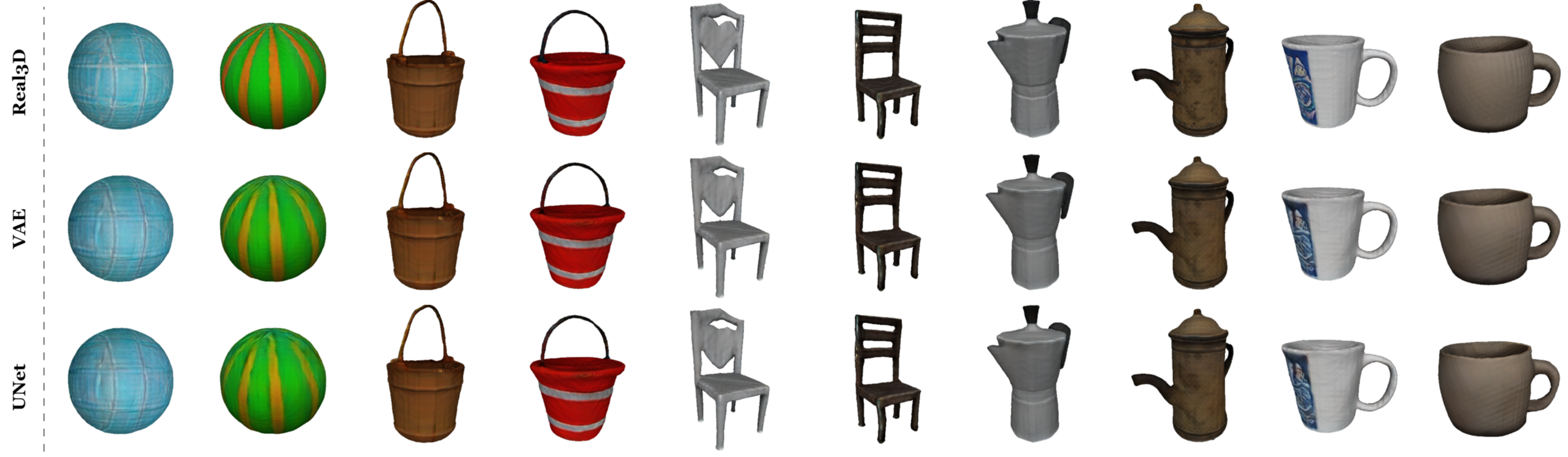}
	\caption{\textbf{Comparison of Meshes Generated by Different Models.} The first row displays the meshes generated by the Real3D LRM model, illustrating its base performance. The second row shows the results from a triplane VAE-based approach, where three separate VAEs were trained for each of the 3 planes of the triplane. The third column presents the meshes produced by the UNet model, which was initially trained with null conditioning. This comparison highlights the varying degrees of control and detail achieved by each method.}
	\label{fig:network_output}
	\vspace{-5mm}
\end{figure*}

\subsection{3D Reconstruction}
\noindent
\textbf{Single-View 3D Reconstruction.} 
Reconstructing the 3D shape of an object from a single image is a challenging problem. Over the years different representations such as voxels~\cite{choy20163d,tulsiani2017multi}, point clouds~\cite{fan2017point,jiang2018gal}, multiplane images ~\cite{mildenhall2019local,tucker2020single}, meshes~\cite{gkioxari2019mesh,liu2019soft}, radiance fields~\cite{rematas2021sharf,yu2021pixelnerf,mildenhall2021nerf, chan2021pi,xu2022point}, signed-distance-functions(SDFs)~\cite{mescheder2019occupancy, park2019deepsdf, sitzmann2019scene}, and 3D Gaussians~\cite{kerbl20233d,szymanowicz2024splatter,tang2024lgm} have been utilized for this challenging task. 
\noindent
Recently, a lot of methods~\cite{poole2022dreamfusion, yi2024gaussiandreamer, tang2024dreamgaussian, jain2021dreamfields} use pre-trained language/image models~\cite{li2022blip, li2023blip, saharia2022photorealistic, rombach2022high} for multi-view guidance for image-to-3D reconstruction. Zero-1-to-3~\cite{liu2023zero1to3} finetunes a Stable-Diffusion model to generate novel views by conditioning with the input images and camera poses. However, maintaining consistency in geometry and colors for the generated images remains a challenge. SyncDreamer~\cite{liu2023syncdreamer} mitigates this by enabling the generation of multi-view consistent images.

\noindent
\textbf{Large Reconstruction Models.} 
Extending pre-trained 2D diffusion models to 3D representation requires test-time optimization i.e., for each text prompt or an input image, an optimization process needs to be executed to produce the output. To mitigate this LRMs~\cite{hong2024lrm} were proposed for generalizable and fast feed-forward 3D reconstruction which follow design principles similar to other foundation models. LRM uses scalable model architecture for encoding diverse shape and texture priors and directly maps 2D information to a 3D representation called the triplane representation. These models are trained with multi-view rendering losses. LRM uses triplane tokens to get information from 2D image features. Other works improve the quality and geometry of the reconstruction through various modifications. TripoSR ~\cite{tochilkin2024triposr} uses the base  LRM model with a series of improvements in data curation, model and training strategy.

CRM~\cite{wang2024crm}, on the other hand, is a transformer-based architecture that does not leverage the geometric priors of the triplane component in its architecture. Instead, it uses a multiview diffusion model to generate 6 orthographic and canonical coordinate maps from a single input image which improves the geometry of the 3D object generated. GRM~\cite{xu2024grm} is an efficient feed-forward 3D generative model that builds on 3DGS, it uses four input images to infer the underlying 3D Gaussians efficiently and can generate 3D from text as well as a single image.

\section{\methodname{}}
\label{sec:motivation}

\noindent
~\methodname{} integrates 3D object generation and text-prompt based editing into a single LRM architecture. With the recent success of diffusion models~\cite{ho2020denoising}, we propose a framework that facilitates conditioning the features in the triplane latent space using user text prompts that enable fine-grained edits into the generation process.
	
\subsection{Training Procedure}
\label{sec:method}

\noindent
Large reconstruction models, typically comprising millions of parameters, are computationally expensive to train from scratch and require a vast dataset of image-to-3D object pairs.
As such, our training procedure only learns the model parameters of the triplane diffusion adapter. We reuse the parameters and architecture of a LRM, namely Real3D~\cite{jiang2024real3d}, to encode the input image into its triplane latent representation, as well as to decode the triplane representation obtained from the triplane diffusion adapter into the edited 3D object. As Real3D has been trained on Objaverse~\cite{deitke2023objaverse}, a large-scale, diverse 3D asset dataset, we expect the parameters of the pre-trained model to generalize well to a semantically diverse set of 3D objects.

\noindent
The triplane diffusion adapter comprises a Tri-VAE (Sec.\ref{subsec:tri_vae}), a variational autoencoder and a conditional diffusion model (Sec.\ref{subsec:diffusion_model}). Tri-VAE compresses the input triplane feature space to train a conditional diffusion model (Sec.\ref{subsec:diffusion_model}), which performs a diffusion process in the triplane latent space. We train the triplane adapter in two phases. In the first phase, we train Tri-VAE to \textit{\textbf{(1)}} encode the input triplane feature space into the latent space corresponding to the input of the conditional diffusion model and \textit{\textbf{(2)}} decode the output of the diffusion model back into the triplane feature space. In the second phase, we freeze the weights of Tri-VAE and train a conditional diffusion model on the compressed triplane latent space to incorporate user-defined edits. Conditioning the latent diffusion process with a text prompt enables the incorporation of edits directly into the latent space, which is then reflected in the generated 3D object. Fig.~\ref{fig:architecture} shows an overview of \methodname{}. 

\subsection{Dataset Preparation}
\label{subsec:dataset_prep}
\vspace{-3mm}
% We aim to utilize the semantically rich triplane latent representation of LRM models and train a diffusion model to edit this triplane latent representation using text-prompts. 
\noindent
To prepare our dataset, we use the popular Objaverse~\cite{deitke2022objaverseuniverseannotated3d} dataset, a large-scale dataset consisting of $800k$ 3D objects spanning diverse categories.

\noindent
\textbf{Dataset for Tri-VAE} 
To train the Tri-VAE, we create a dataset of images of 3D objects and their triplane features through a pre-trained LRM model. To this end, we first sample $12584$ 3D objects from the curated Objaverse LVIS dataset. Subsequently, we render different views of the object by sampling a random pose from a set of manually selected views around the object. $1024\times1024$ image from the selected pose is then rendered and passed through a pre-trained LRM model $\mathbb{M}$ to get the triplane features $x$, where $x \in \mathcal{R}^{ H_1 \times W_1 \times C_1 \times 3}$. We refer to this dataset as the triplane feature dataset ($\mathbb{D}_{base})$.

\noindent
\textbf{Dataset for the Conditional Diffusion Model}
To prepare a dataset to train the conditional diffusion model, we first generate a dataset of image pairs for $2108$ objects, where each pair consists of a rendered image $\mathcal{I}$ of an object and an edited image  $\mathcal{I}^{edit}$ of the same object edited based on the edit text-prompt $e$. We utilize InstructPix2Pix~\cite{brooks2023instructpix2pix}, a conditional image-to-image diffusion model, to generate the edited images in this dataset. We manually filter these pairs after generation to retain only high-quality, accurately edited images due to the varying quality of the edited images from InstructPix2Pix. We obtain the triplane features of the original and edited images, $x$ and $x^{edit}$, respectively, by passing $\mathcal{I}$ and $\mathcal{I}^{edit}$ through the pre-trained LRM model $\mathbb{M}$ to create a paired triplane dataset with edit prompts as $\mathbb{D}_{paired}= \{\left( \left(x_i,x^{edit}_i\right),e_i\right) | i\in M\}$, where $M$ is the size of this dataset. Additionally, we also create another dataset with null edit prompts for each object, to signify that no edit should be done to the latent representation $\mathbb{D}_{identity}= \{\left( \left(x_i,x_i\right),``."\right) | i\in N\}$, where $N$ is the size of such examples. We train with $\mathbb{D}_{identity}$ to preserve the geometry details in the final output. 

\subsection{Tri-VAE: Triplane Variational Auto-Encoder}
\label{subsec:tri_vae}
\noindent
LDM~\cite{rombach2022high} based diffusion models apply the diffusion process in the latent space of the image encoder instead of the pixel space as it reduces the computational requirements of the training procedure. Additionally, this latent space is usually semantically rich, which is leveraged by different works~\cite{brooks2023instructpix2pix,chen2024anydoor,zhuang2023task,ruiz2023dreambooth} for the editing tasks. In this work, we train a VAE to operate on the triplane latent features of the LRM model, which, to the best of our knowledge, has not been done before. 

\noindent
Our Tri-VAE architecture includes a 2D image autoencoder and is trained using a combination of traditional reconstruction loss and KL divergence loss. Each plane of the input triplane feature is passed independently to Tri-VAE. More specifically, given an input triplane feature $x \in \mathcal{R}^{ H_1 \times W_1 \times C_1 \times 3}$ from the triplane-feature extractor, we first extract each plane of the input triplane feature $x_k \in \mathcal{R}^{ H_1 \times W_1 \times C_1 }$ where $k\in\{1,2,3\}$. Then, encoder $\mathcal{E}$ of Tri-VAE encodes $x_k$ into a latent representation $z_k = \mathcal{E}\left(x_k\right)$ where $z_k \in \mathcal{R}^{ H_2 \times W_2 \times C_2 }$. Subsequently, decoder $\mathcal{D}$ reconstructs back the plane of the triplane, giving $\tilde{x_k} = \mathcal{D}\left(z_k\right)$. We then concatentate the three reconstructed planes, $\tilde{x} = \left[\tilde{x_1},\tilde{x_2},\tilde{x_3}\right]$ where $\left[.\right]$ is the concatenation operator, to get the reconstructed triplane. We train Tri-VAE with following loss:
\begin{equation}
	L_{VAE} = \frac{1}{N}\sum_{i=1}^{N}\left(x_k - \tilde{x_k}\right)^2  + \lambda_{KL}L_{KL} 
\end{equation}

%We provide the implementation details for Tri-VAE in Sec.~\ref{subsec:implementation_details}.

\noindent
\textbf{Training of Tri-VAE.} 
We train Tri-VAE autoregressively on dataset $\mathbb{D}_{base}$ described in Sec.~\ref{subsec:dataset_prep}. We normalize these triplane features in the range of $\left[-1,1\right]$ by calculating the minimum and maximum for each plane on the entire dataset to stabilize the training process. Normalization of triplane latent features is as follows:
\begin{equation}
	\hat{x_k} = \frac{x_k - x^{min}_k}{x^{max}_k - x^{min}_k}; \;\; k=\{1,2,3\}
\end{equation}  
where $x^{min}_k$ and $x^{max}_k$ are minimum and maximum of the input triplane features on the entire dataset.

%After normalization the results were still not good enough, it was due to the high number of channels in each plane of the triplane (80 per plane in case of triplane generated by OpenLRM) which the Unet \cite{ronneberger2015unet} was not able to learn. Hence, we thought of training 3 separate VAE autoencoders \cite{kingma2022autoencoding} for each plane which encodes the channels in each plane of the triplane to 4 channels. Then the normalized planes of triplanes were feed to the VAEs for training. We used a latent space dimension of C = 4 and plane resolution of H = W = 16, while training of the VAEs we have used LeakyReLU activation function since there some negative values in the planes of the triplane. Architecture of our VAEs has three down-sample and upsample blocks with 3 layers per block and block out channels =[128,256,512]. In total each VAE has around 68M parameters and is trained with AdamW optimizer and linear learning rate decay. We train the VAEs using a A6000 GPU for 16 hours. To see to that our adapter works with other LRM models as well we have tested it for another model that is Real3D. For the adapter to work all the experiments should be repeated with triplanes generated by Real3D.
% we train a diffusion model to learn the distribution of
\subsection{LTriD: Latent TriPlane Diffusion Model}
\label{subsec:diffusion_model}
\noindent
Similar to the LDM~\cite{rombach2022high} based models, we train our diffusion model on the latent space of the encoder $\mathcal{E}$ of TriVAE (Sec.~\ref{subsec:tri_vae}). LDMs use cross-attention layers for conditioning using text-prompts or other conditional input for the controlled generation and editing tasks. We also employ these cross-attention layers to provide text-based edit instructions. We train our diffusion model in two ways: \textbf{\textit{1.)}} Train with null text prompts to perform no edits, and \textbf{\textit{2.)}} Train with edit prompts to edit the triplane representation.

\noindent
Specifically, for each plane $x_k$ in the input triplane $x$, encoder $\mathcal{E}$ encodes it into a latent representation $z_k$, where $z_k \in \mathcal{R}^{ H_2 \times W_2 \times C_2 }$. We stack latent representations of each triplane into a single latent representation $z = \left(z_1,z_2,z_3\right)$, where $\left(.\right)$ is the stack operation and $z \in \mathcal{R}^{ H_2 \times W_2 \times 3C_2 }$. Subsequently, we pass $z$ through a diffusion process by progressively adding noise $\epsilon_t$ to get the noisy latent $z_t$ for timestamps $t\in \{1..T\}$. Further, to preserve the original structure of the 3D representation, we concatenate the original latent $z$ to the noisy latent and learn a network $\epsilon_{\theta}$ by minimizing the following loss function. $\epsilon_{\theta}$ predicts the noise added to the noisy latent $z_t$ given text-conditioning $c_T$.

\begin{equation}
	L_{LTriD} = \;\; E_{x_0,\epsilon \sim \mathcal{N} \left( 0,I \right) ,t} ||\epsilon -\epsilon _{\theta}\left(\left[z_t,z\right], t, c\right)||^2 
	\label{eq:diff_loss}
\end{equation}   
where $\left[.\right]$ is the concat operator.

\noindent
In order to preserve the geometric details of the 3D objects, we first train the denoiser network  $\epsilon_{\theta}$ with $\mathbb{D}_{identity}$ to learn the distribution of the triplane features of the original 3D objects. Subsequently, we train it with the paired data $\mathbb{D}_{paired}$ described in Sec.~\ref{subsec:dataset_prep} to learn to edit the triplane representation with text prompts.

\vspace{-1.5mm}
\subsection{Inference}
\noindent
During inference, we provide an image of the object and the edit text-prompt to our method. The edited 3D object from a single image can be generated by first passing the image through the pre-trained LRM encoder to obtain the triplane latent features, then generating the triplane latent representation of the edited 3D object using the triplane diffusion adapter model and finally passing these generated triplane latent features through the pre-trained LRM decoder.

% Summary : 

% - InstructTri2Tri -> triplane latents
% - triplane adapter : 
%     - latents -> vae -> diffusion process
% - add text prompt embeddings in diffusion process
% - decode using vae to new triplane latents of the edited 3d model
% - use lrm to decode triplane latents to object model

% Triplane adapter

% Training : 
%  - Dataset generation
%  Phase 1 : 
%     VAE Training
%     - Losses : VAE loss

%  Phase 2 : 
%     Diffusion model Training
%     -  diffusion model loss

% We also use the LeakyReLU activation function since incorporating non-zero slope for the negative part in rectified activations can improve the results~\cite{xu2015empirical}.
\subsection{Implementation Details}
\label{subsec:implementation_details}
\noindent
\textbf{Tri-VAE} 
We use Real3D~\cite{jiang2024real3d} as the base LRM model for~\methodname{}. Real3D computes the triplane features with dimensions $H_1=64$, $W_1=64$ and $C_1=40$ (for each plane of the triplane) given an image. We encode these planes of the triplane features into a latent space and concatenate the channels to obtain a triplane latent of dimensions $H_2=16$, $W_2=16$ and $C_2=12$, which is further given to the diffusion model. Each VAE has around $68M$ parameters and is trained with AdamW~\cite{adamw} optimizer, linear learning rate decay with a learning rate of $5e^{-5}$ and LeakyReLU activation function. We train the VAEs using an A6000 GPU for 16 hours with a batch size of $16$ for $256$ epochs.

\noindent
\textbf{Latent TriPlane Diffusion Model.} We provide the architecture of the denoiser network $\epsilon_{\theta}$ in the supplementary material. $\epsilon_{\theta}$ has around 870M parameters and we train it with AdamW~\cite{adamw} optimizer, linear learning rate decay with a learning rate of $5e^{-5}$ and LeakyReLU activation function. Using DDPM Scheduler~\cite{ho2020denoising} we add noise to the triplane latent obtained from Tri-VAE over $1000$ timesteps .This model was trained for one day on an A6000 GPU with a batch size of $8$ for $25000$ steps. During inference we generate the edited triplane latent with $100$ denoising steps using an Euler ancestral sampler with denoising variance schedule given by~\cite{karras2022elucidating}.

\begin{figure*}[h]
	\centering
	\includegraphics[width=\linewidth]{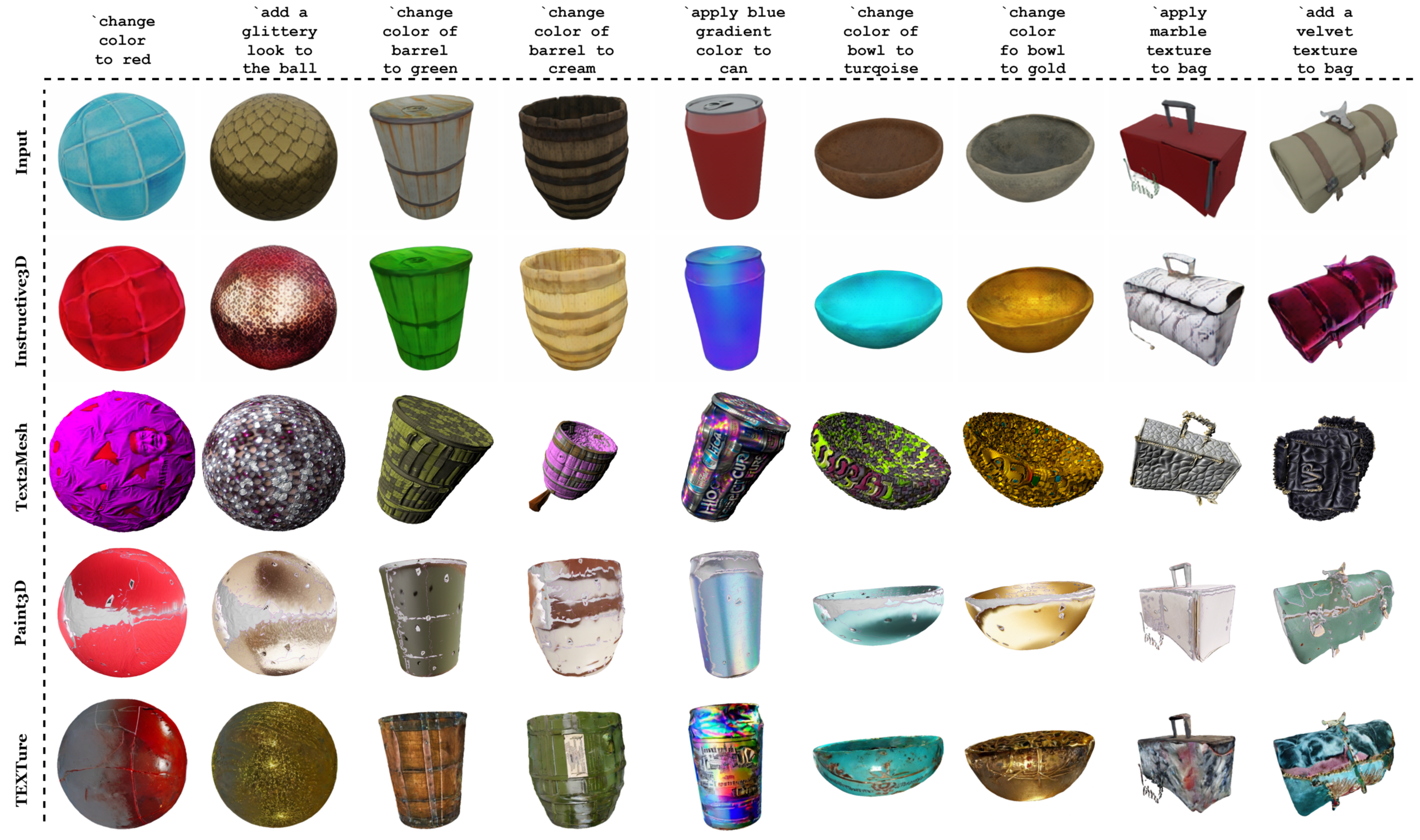}
	\caption{\textbf{Comparison of Text-Conditioned 3D Generation between \methodname{}(Ours) and baselines.} The first row lists the input text prompts. The second row displays the input image, third row show results for our \methodname{} and subsequent rows show results from other baselines. This comparison highlights the effectiveness of our method \methodname{} in providing fine-grained control with the input text descriptions. Notice how other baselines fail to produce edit for the text prompt \textit{``add a velvet texture to bag''}(last column)}
	\label{fig:t2m}
	\vspace{-5mm}
\end{figure*}
\section{Experiments}

\subsection{Evaluation}
\noindent
\textbf{Datasets.}We use the Objaverse LVIS annotations dataset for both training and evaluation. We explain the data preparation in Sec.~\ref{subsec:dataset_prep}. For evaluation we also generate a synthetic dataset of $45$ objects, capturing $14$ images per object from different viewpoints. These views correspond to $8$ vertices and $6$ face directions of a cube, providing comprehensive coverage of each 3D model.

\noindent
\textbf{Baselines.}We compare our method with Text2Mesh~\cite{Michel2021Text2MeshTN}, Paint3D~\cite{zeng2023paint3dpaint3dlightingless} and TEXTure~\cite{richardson2023texturetextguidedtexturing3d}, which takes a 3D mesh and a text prompt as input and generates an output mesh with the given text conditioning. We provide the mesh generated by Real3D to these 3 models with an edit prompt and compare the output with our generated mesh.

\noindent
\textbf{Metrics.} We evaluate the objects using a combination of perceptual, geometric, and alignment-based metrics:
\begin{itemize}
	\vspace{-3mm}
	\item \textbf{Perceptual Quality Metrics:} We compute PSNR, SSIM, and LPIPS~\cite{zhang2018unreasonable} between the prediction and the ground truth. This captures how close the visual appearance of the edited 3D mesh is to the target mesh. Further, we also compute the FID distance~\cite{heusel2017gans} which evaluates the quality of realism of the edited 3D mesh. We also evaluate KID~\cite{bińkowski2021demystifyingmmdgans} which is more robust on smaller sample sizes. 
	\vspace{-3mm}
	\item \textbf{Text Alignment Metric:} We use 2D CLIP~\cite{radford2021learning} score to evaluate the correlation between the caption for an image and the actual content.
\end{itemize}
%Similar to the evaluation protocols in InstructNeRF2NeRF~\cite{Haque2023InstructNeRF2NeRFE3}, we measure alignment of the performed 3D edit with the edit instruction using CLIP Direction consistency score. It measures the directional similarity between pairs of original and edited images in adjacent frames of novel rendered camera paths. We render $10$ views around the mesh.    
%For quantitative comparions we use OpenLRM, we pass the after edit image generated by InstructPix2Pix to OpenLRM, then generate the 3D object and then compare it with the 3D object generated through our adapter. We first render views from our objects and then use CLIP 2D score and CLIP directional similarity score to show quantitative results.

\subsection{Experimental Results}

\noindent
\textbf{Qualitative Results.}
We present qualitative results from our method across various categories of objects, illustrating the strength of our approach in generating fine-grained 3D edits aligned with text prompts. Fig.~\ref{fig:teaser} shows the limitations of current LRM models and how our method improves control over the generated 3D objects. The top row illustrates the lack of fine-grained control in existing models, while the bottom row highlights the precision and adaptability of our method through several edited 3D objects.

\noindent
Fig.~\ref{fig:network_output} compares the meshes generated by the VAEs and the UNet in our method. The first row shows the results from the Real3D LRM model. The second and the third rows show the meshes generated by our models, which demonstrates the superior geometry preservation and smoother transformations. Fig.~\ref{fig:t2m} compares our method with Text2Mesh, Paint3D, and TEXTure. The first row shows the input mesh, while the second row displays the output from our method. Subsequent rows show the mesh outputs produced by baseline methods when given a pre-edit object from Real3D and the corresponding text prompt. Our approach effectively applies edits such as adding texture or color without distorting the original geometry, outperforming other baselines in preserving the geometry.

\begin{figure*}
	\centering
	\includegraphics[width=0.825\linewidth]{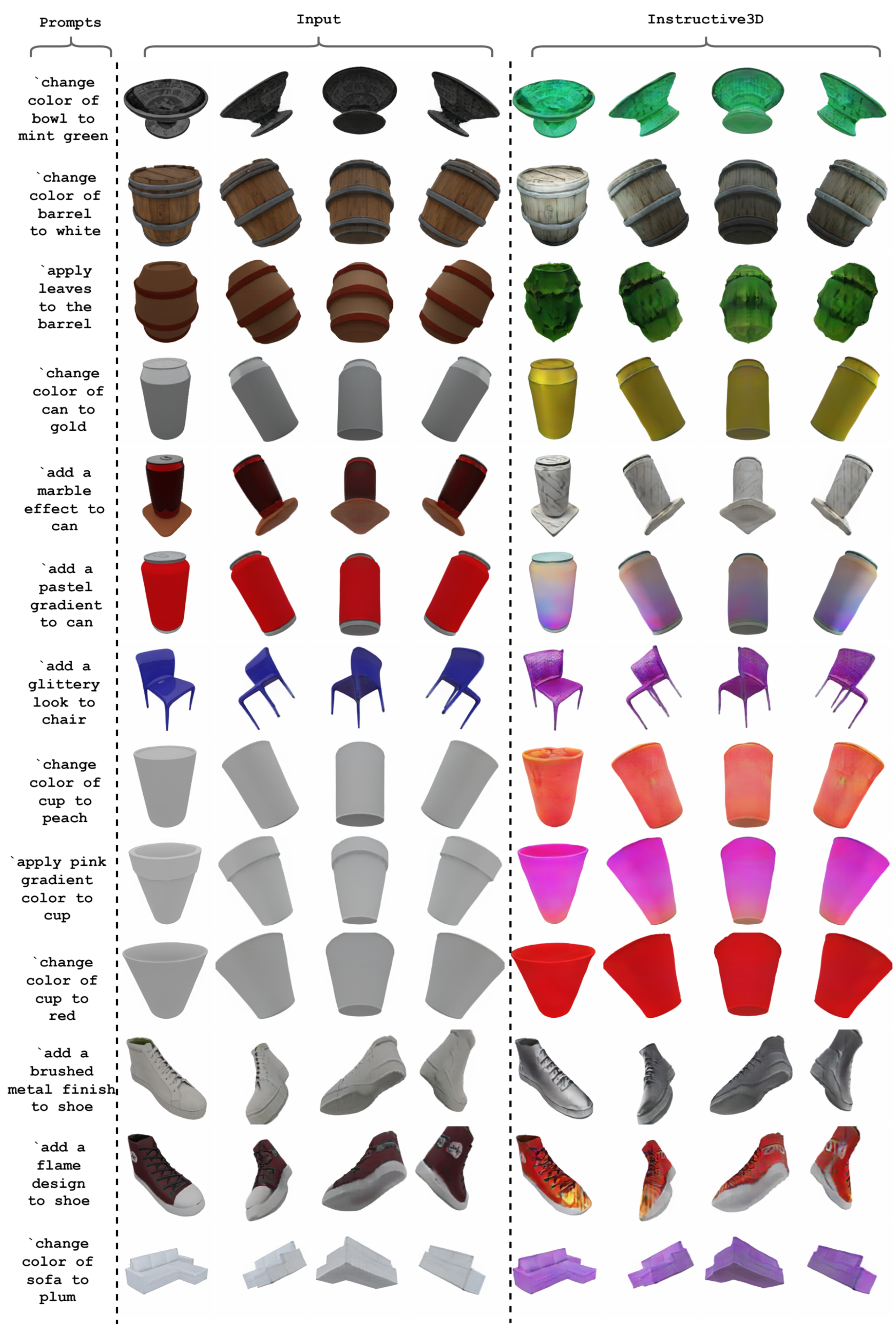}
	\caption{\textbf{Additional Results of our adapter ~\methodname{} using Real3D.} The first column shows the text prompts used for their respective objects. The second column shows four input images of a single object captured from different angles. The third column presents the corresponding outputs generated using our ~\methodname{} method, demonstrating how the model adapts to the object based on text prompts. These results highlight the enhanced control offered by our method.}
	\label{fig:more_results}
\end{figure*}

\noindent
Fig.~\ref{fig:more_results} provides more results using our method. We show results from $4$ different angles. We observe that the edits are aligned with the input text prompts and preserve the object's geometry and texture across different views. These visualizations confirm that our method excels in producing high-quality, text-conditioned 3D edits, preserving geometric integrity while responding effectively to the given prompts.

% \usepackage{booktabs}
\begin{table}
	\centering
	\caption{\textbf{Quantitative Comparison of~\methodname{} with baselines.} Our method outperforms the baselines in almost all metrics, demonstrating superior performance.}
	\resizebox{0.5\textwidth}{!}{
		\begin{tabular}{ccccccc}
			\toprule
			 & \textbf{CLIP}~\cite{radford2021learning} $\uparrow$ & \textbf{LPIPS}~\cite{zhang2018unreasonable} $\downarrow$ & \textbf{SSIM} $\uparrow$ & \textbf{FID}~\cite{heusel2017gans}  $\downarrow$ & \textbf{KID}~\cite{bińkowski2021demystifyingmmdgans} $\downarrow$ & \textbf{PSNR} $\uparrow$ \\
			\midrule
			\textbf{Text2Mesh} & 21.86 & 0.1986 & 0.858 & 273.12 & 0.1785 & 37.25  \\
			\textbf{Paint3D} & 21.78 & 0.1061 & 0.931 & 209.87 & 0.1463 & 39.48  \\
			\textbf{TEXTure} & \textbf{21.98} & 0.0995 & 0.929 & 222.04 & 0.1311 & 39.54  \\
			\textbf{~\methodname{}(Ours)} & 21.94 & \textbf{0.0278} & \textbf{0.951} & \textbf{114.91} & \textbf{0.0457} & \textbf{40.47} \\
			\bottomrule
		\end{tabular}
	}
	\label{tab:Metrics2}
\end{table}

\begin{figure}[!t]
	\centering
	\includegraphics[width=\linewidth]{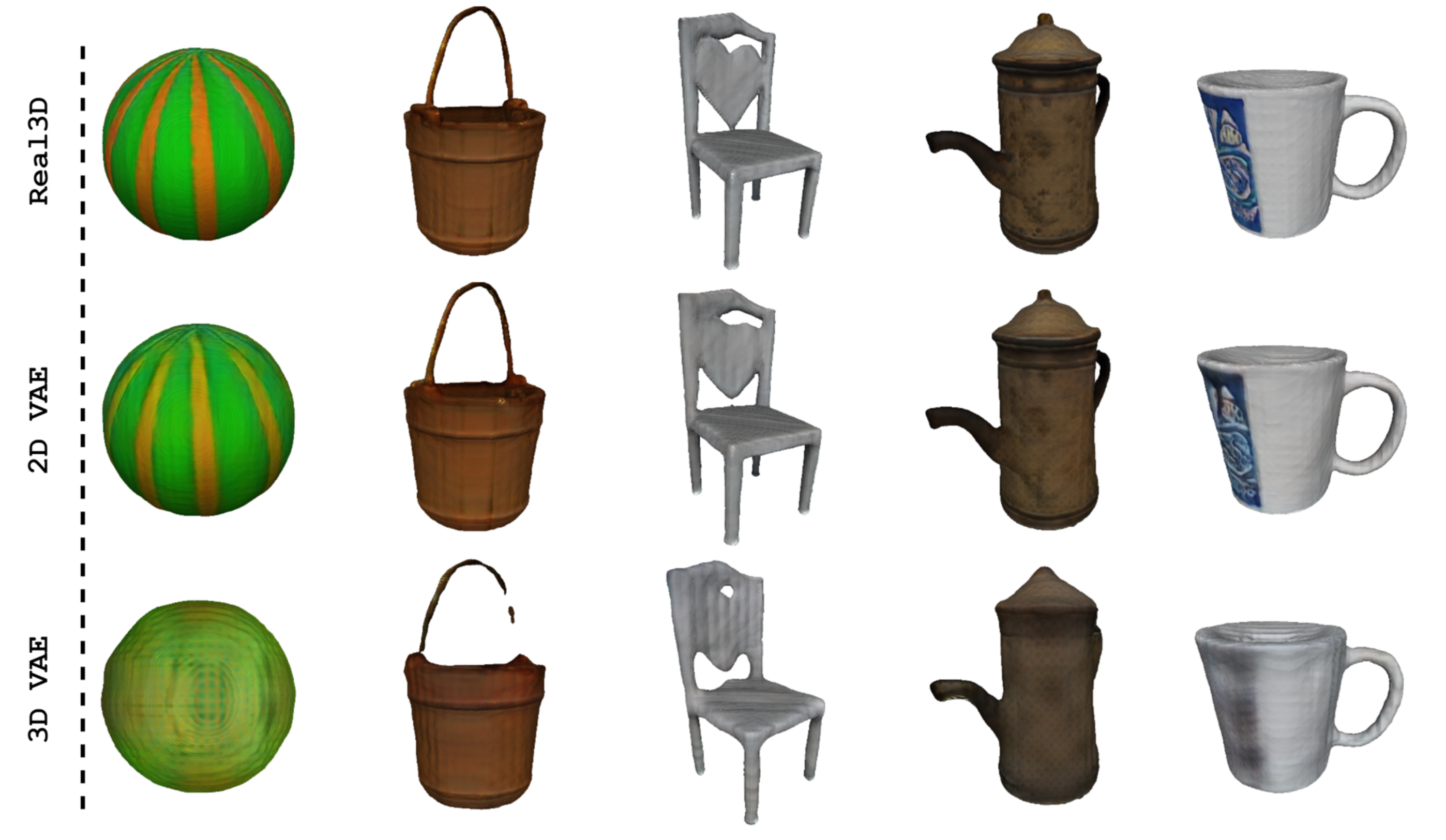}
	\caption{\textbf{Comparison of 2D and 3D VAE.} The first row shows the output generated by Real3D and the next two rows compare the output generated by 2D VAE and 3D VAE respectively.}
	\label{fig:ablation}
	\vspace{-7mm}
\end{figure}

\noindent
\textbf{Quantitative Results} We calculate the metrics by using the after-edit object generated by Real3D and compare it with the objects generated by our method and other baseline methods. As shown in Tab.~\ref{tab:Metrics2}, our method consistently outperforms prior works across all key metrics including LPIPS, PSNR, SSIM, FID, KID, and CLIP scores. Specifically, our approach demonstrates superior performance in both perceptual quality and text alignment. We used a dataset of $45$ objects, capturing $14$ images per object from various viewpoints to calculate these metrics, including all $8$ vertices and $6$ face directions of a cube.

\noindent
Our method showcases substantial improvements in LPIPS and PSNR, which reflect better perceptual similarity and reconstruction fidelity. Moreover, the lower FID and KID scores indicate that our generated outputs are more realistic and aligned with the natural image distribution compared to the baseline methods. The high CLIP score validates that our edits are semantically closer to the text prompt. This highlights the effectiveness of our approach in producing accurate and high-quality 3D edits.

%We show quanitative results in Tab.~\ref{tab:Metrics2}. Our method outperforms Text2Mesh, Paint3D and TEXTure across all metrics. We calculate the CLIP score using the after edit objectWe calculate these metrics by using the initial object generated by the Real3D model as the ground truth and compare that with the objects generated by our method, Text2Mesh, Paint3D and . These results underscore the benefits of editing directly in the triplane representation, which retains rich geometric information and allows for more precise edits.

\subsection{Ablation Studies}
\noindent
We conducted an ablation study to evaluate the design choices in our architecture. In particular, we experimented with using three separate 2D VAEs for each plane of the triplane instead of a single 3D VAE. Each plane of the triplane has a large number of channels, so separating the planes led to minimal information loss when converted to latent space. This design proved effective in maintaining high fidelity while reducing computational complexity. In Tab.~\ref{tab:ablation} we compare the training loss and MSE Loss between input and output for the two different methods and also show it qualitatively in Fig.~\ref{fig:ablation}.

\begin{table}
	\centering
	\caption{\textbf{Ablation Study.} The table presents comparison between 2D and 3D VAE in terms of training loss and MSE Loss between the input and output triplane. We use $100$ objects to compute the loss. We observe that 2D VAE gives superior results.}
	\begin{tabular}{ccc}
		\toprule
		Model & \textbf{Training loss} $\downarrow$ & \textbf{MSE}  $\downarrow$ \\
		\midrule
		3D VAE & $17.9\times10^{-5}$ & $18.43\times10^{-5}$  \\
		2D VAE & $\mathbf{4.19\times10^{-5}}$ &  $\mathbf{11.77\times10^{-5}}$  \\
		\bottomrule
	\end{tabular}
	\label{tab:ablation}
\end{table}

\noindent
\textbf{Limitations}
Since we are working with triplanes there is no ground truth dataset for them. We use the LRM model for generating all our triplanes, so if the triplane generated by this model is not good enough then the output of our method might not be that good.

\section{Conclusion}
\noindent
In this paper, we propose~\methodname{}, a LRM based method for integrating text-based prompts into the generated 3D objects. We achieve this through the introduction of a triplane adapter that performs a latent diffusion process in the triplane latent space conditioned on the edit prompts, thus incorporating the user-based edits into the generation process in a geometrically consistent manner. We show the effectiveness of our approach both quantitatively and qualitatively with different baselines. These results show that our method has successfully integrated the text-conditioning in the LRM models. ~\methodname{} produces much better quality 3D objects that preserve their geometrical properties, which the baseline fails to retain. 

\noindent
Additionally,~\methodname{} is also compute and data efficient, as it uses pre-trained LRM models to extract the triplane latents from images and decode them into 3D objects and thus requires only the triplane diffusion adapter to be trained. The results we showcase are obtained from a model that has been trained only with about 14.5K 3D objects.

%\input{source/X_suppl.tex}
%\clearpage

%%%%%%%%% REFERENCES
{\small
\bibliographystyle{ieee_fullname}
\bibliography{egbib}
}

\clearpage
\appendix

\tableofcontents

\addtocontents{toc}
{\protect\setcounter{tocdepth}{2}}

\section{Introduction}
We present additional results and other details related to our proposed method :~\methodname{}. 
%For ease of access, we provide a static webpage ``Instructive3D.html'' in the attached supplementary material. We explain the orgianization of the attached webpage  ``Instructive3D.html'' in Appendix~\ref{sec:html_details}. 
We present implementation details in Appendix ~\ref{sec:implementation_details}. We present additional experimental results in Appendix \ref{sec:experimental_results}.

\section{Implementation Details}
\label{sec:implementation_details}
For Tri-VAE, we use $3$ DownEncoderBlock2D for the encoder and $3$ UpDecoderBlock2D for the decoder, with $3$ layers per block. The number of in channels and out channels is $40$ for each VAE and the number of channels in latent space is $4$ per plane of the triplane. The sample size used is $64$.

For Latent TriPlane Diffusion model, we use $3$ CrossAttnDownBlock2D along with $1$ DownBlock2D for the encoder part of the model and $3$ CrossAttnUpBlock2D along with $1$ UpBlock2D for the decoder part, also we use $1$ UNetMidBlock2DCrossAttn in the middle, the text embedding obtained from the CLIP~\cite{radford2021learning} transformer is fed to all the cross attention blocks. We use $2$ layers per block, the number of in channels is $24$ and out channels is $12$, with a sample size of $16$. In the background the UNetMidBlock2DCrossAttn, CrossAttnDownBlock2D and CrossAttnUpBlock2D uses BasicTransformerBlock2D, in which both self attention and cross attention is enabled.

\section{Experimental Results}
\label{sec:experimental_results}
\begin{figure*}
	\centering
	\includegraphics[width=\linewidth]{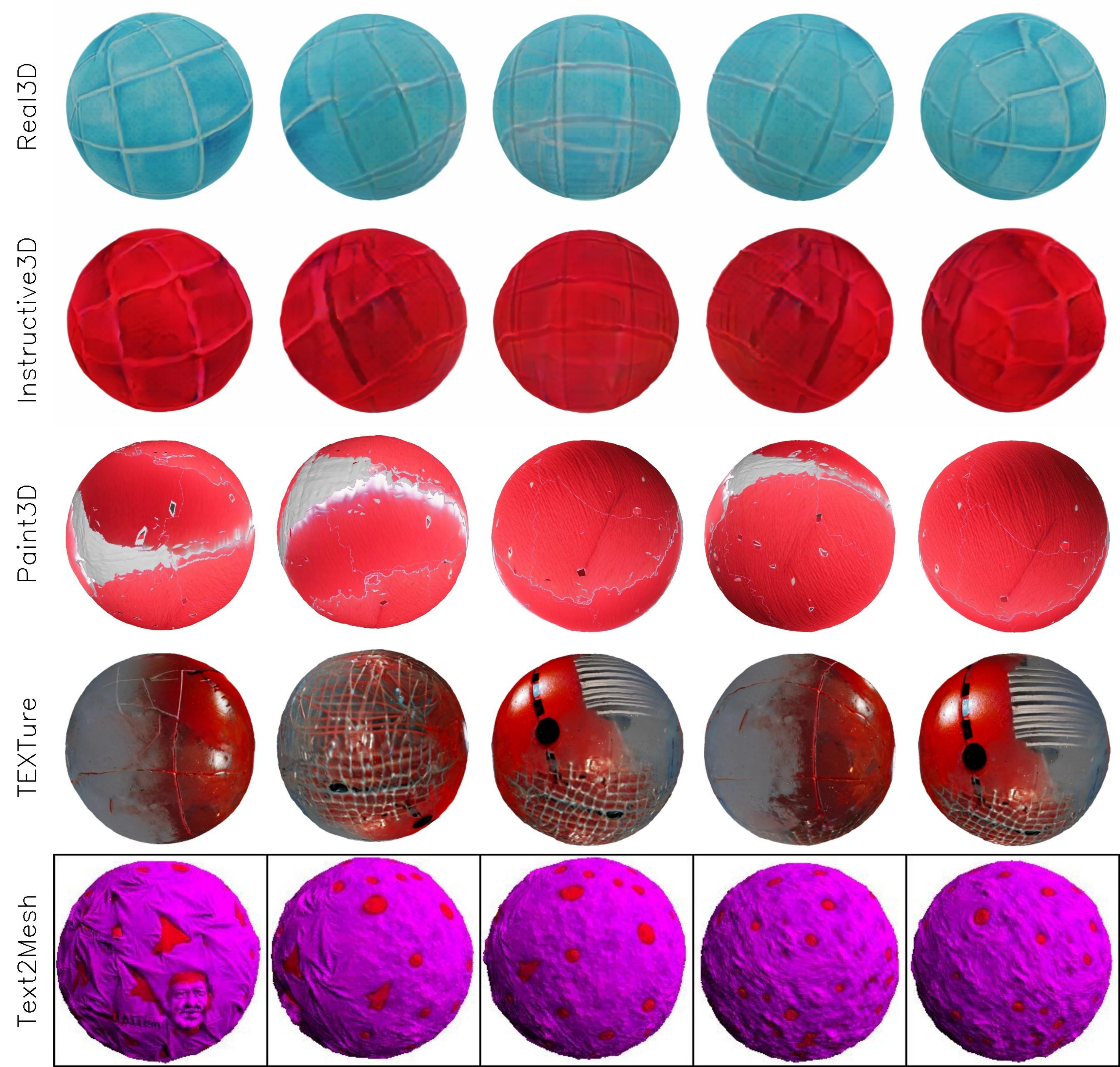}
	\caption{\textbf{Baseline comparison results.} Top row shows the rendered images from the mesh obtained from Real3D~\cite{jiang2024real3d}. Second row shows results from our method. Caption used for editing is: \textit{``change color to red''}.}
	\label{fig:supp_1}
\end{figure*}

\begin{figure*}
	\centering
	\includegraphics[width=\linewidth]{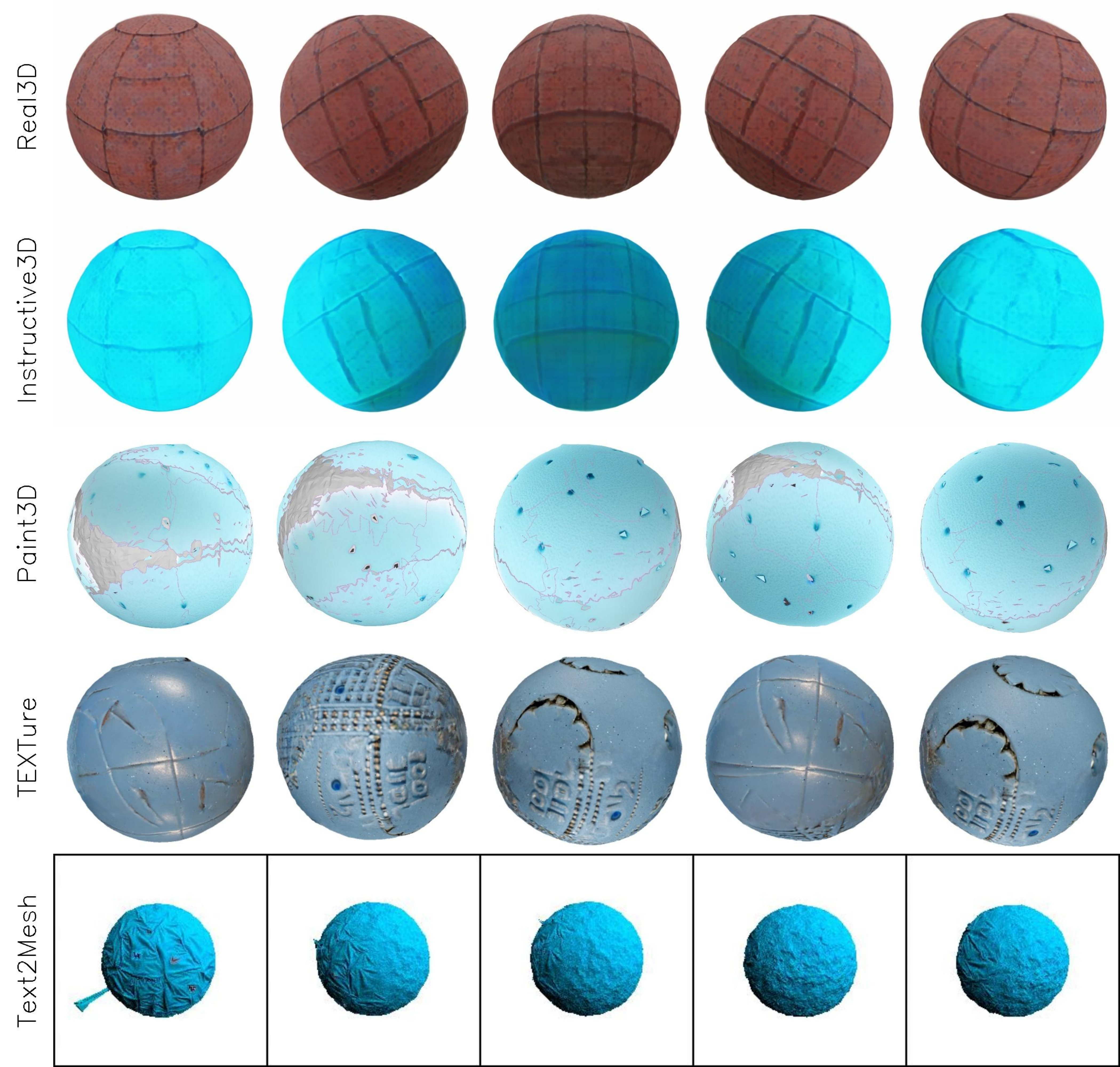}
	\caption{\textbf{Baseline comparison results.} Top row shows the rendered images from the mesh obtained from Real3D~\cite{jiang2024real3d}. Second row shows results from our method. Caption used for editing is: \textit{`change color to powder blue''}.}
	\label{fig:supp_2}
\end{figure*}

\begin{figure*}
	\centering
	\includegraphics[width=\linewidth]{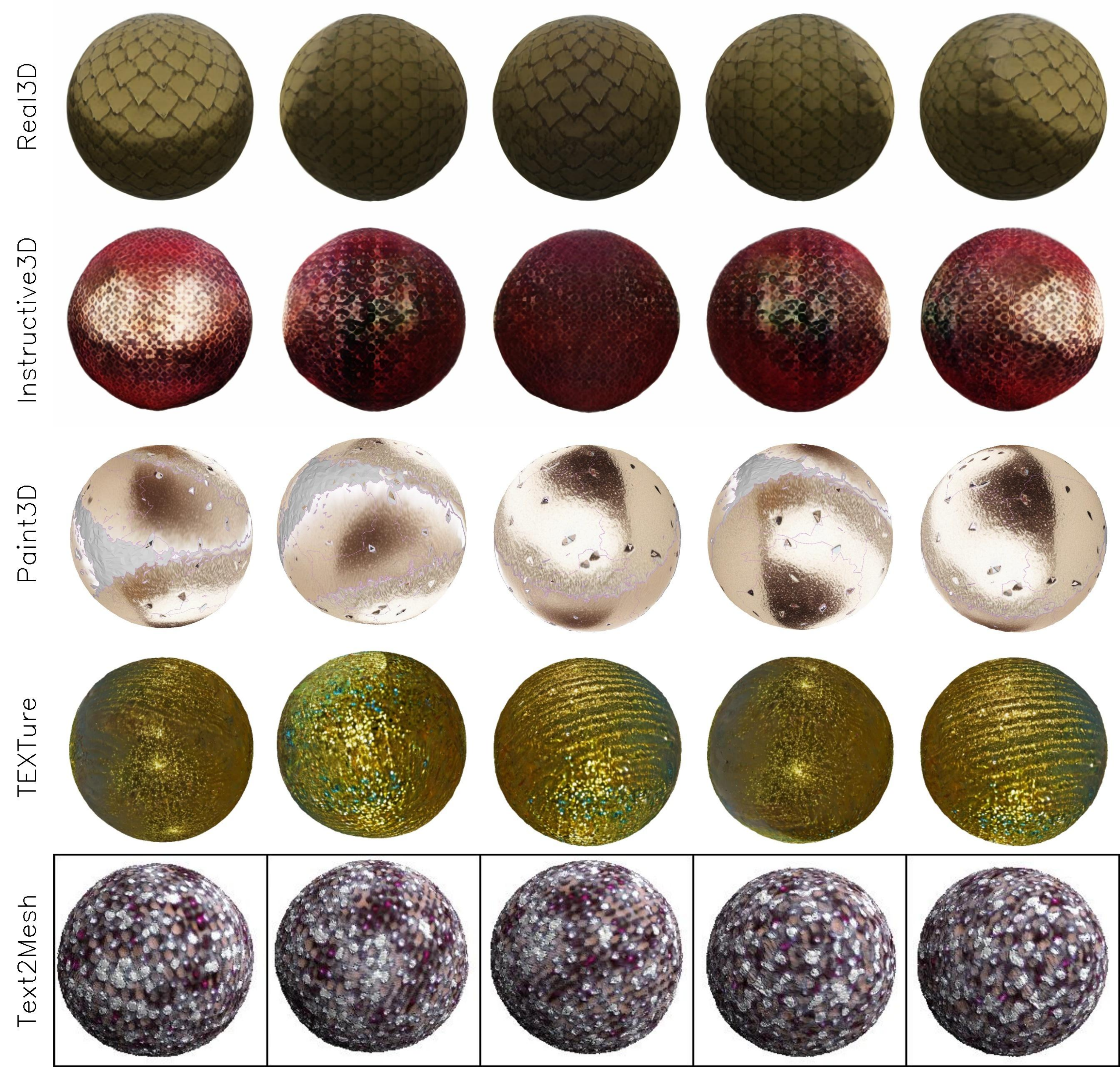}
	\caption{\textbf{Baseline comparison results.} Top row shows the rendered images from the mesh obtained from Real3D~\cite{jiang2024real3d}. Second row shows results from our method. Caption used for editing is: \textit{`add a glittery look to the ball''}.}
	\label{fig:supp_3}
\end{figure*}

\begin{figure*}
	\centering
	\includegraphics[width=\linewidth]{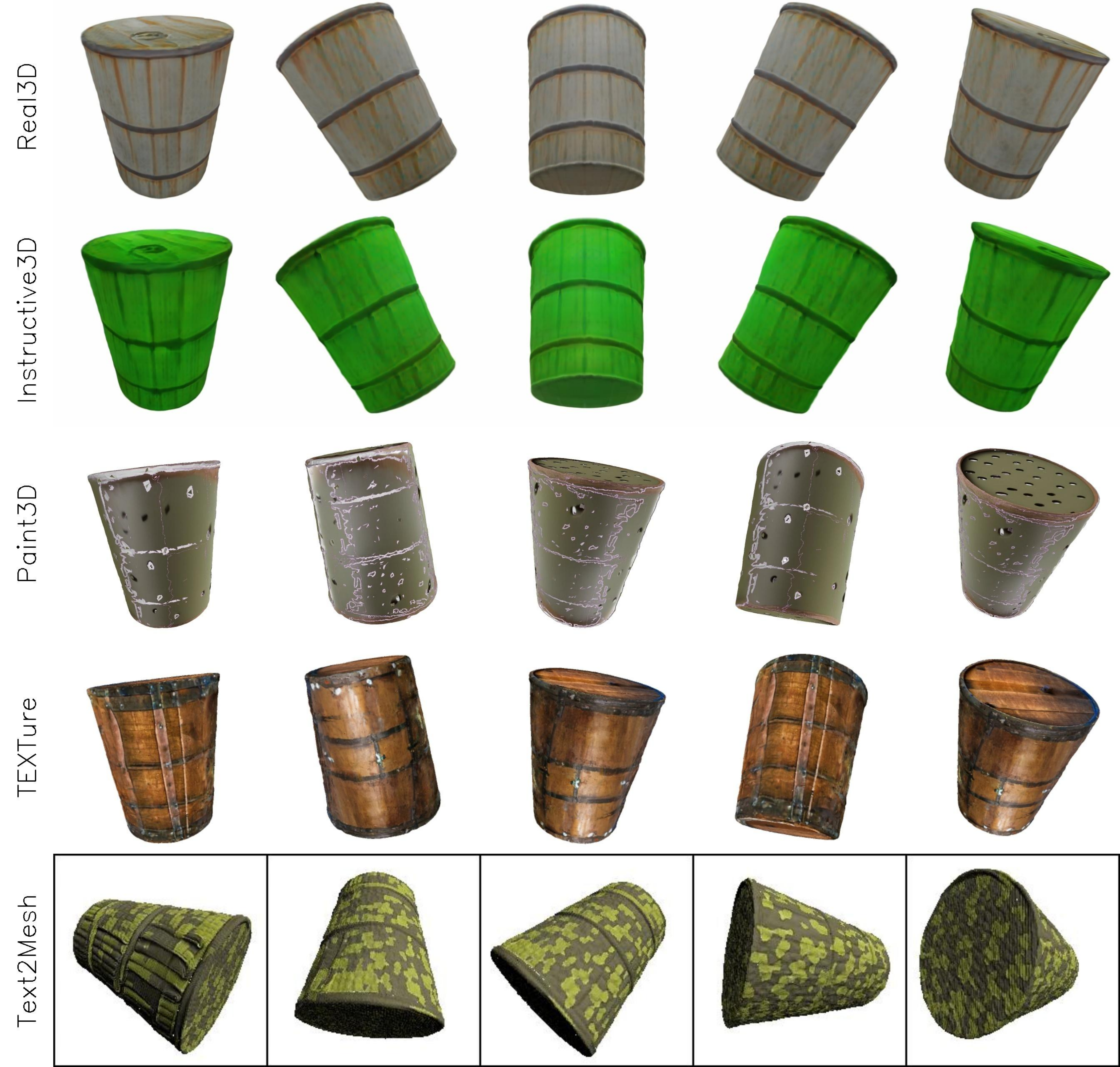}
	\caption{\textbf{Baseline comparison results.} Top row shows the rendered images from the mesh obtained from Real3D~\cite{jiang2024real3d}. Second row shows results from our method. Caption used for editing is: \textit{`change color of barrel to bamboo
			green''}.}
	\label{fig:supp_4}
\end{figure*}

\begin{figure*}
	\centering
	\includegraphics[width=\linewidth]{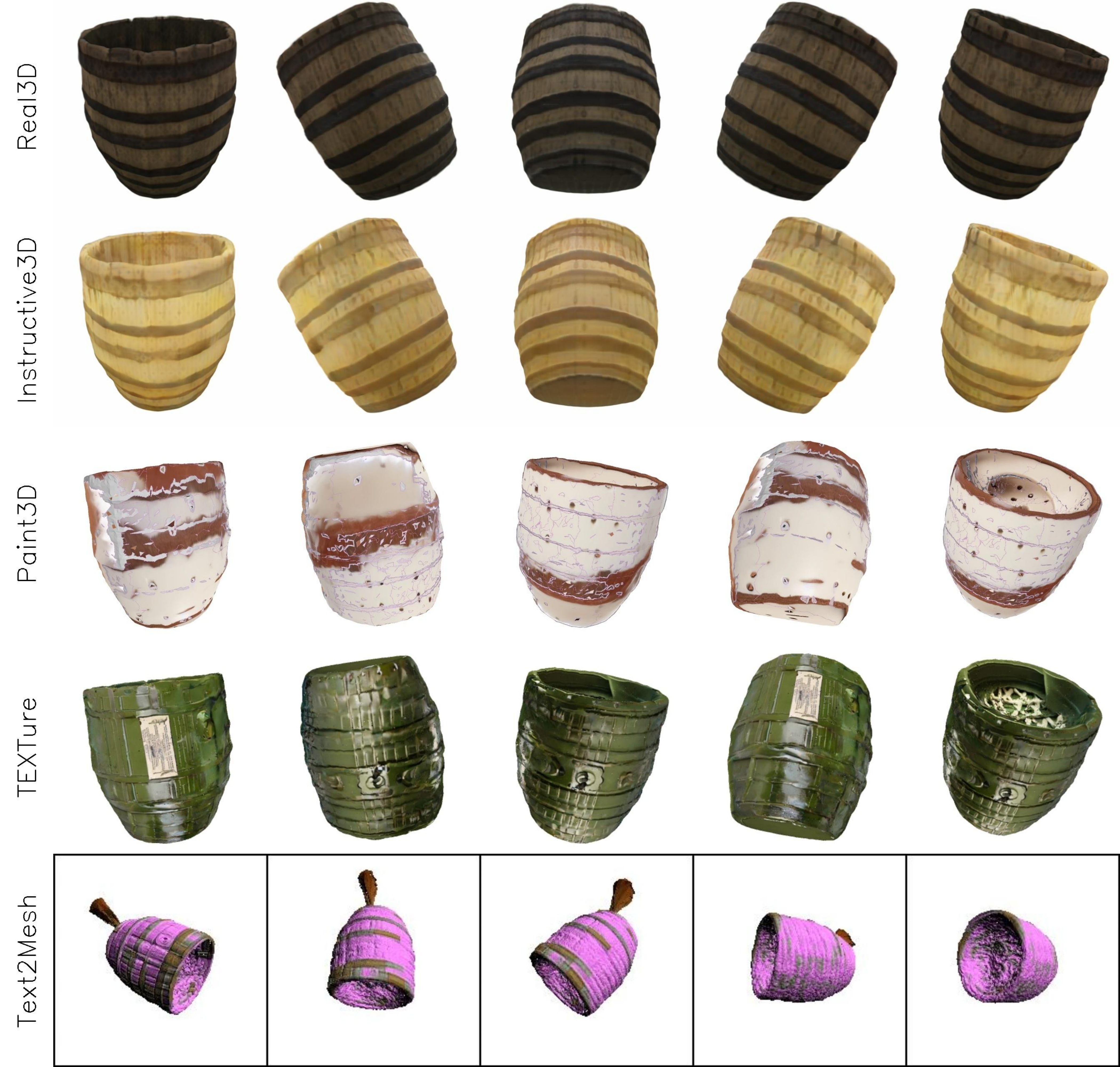}
	\caption{\textbf{Baseline comparison results.} Top row shows the rendered images from the mesh obtained from Real3D~\cite{jiang2024real3d}. Second row shows results from our method. Caption used for editing is: \textit{`change color of barrel to cream''}.}
	\label{fig:supp_5}
\end{figure*}

\begin{figure*}
	\centering
	\includegraphics[width=\linewidth]{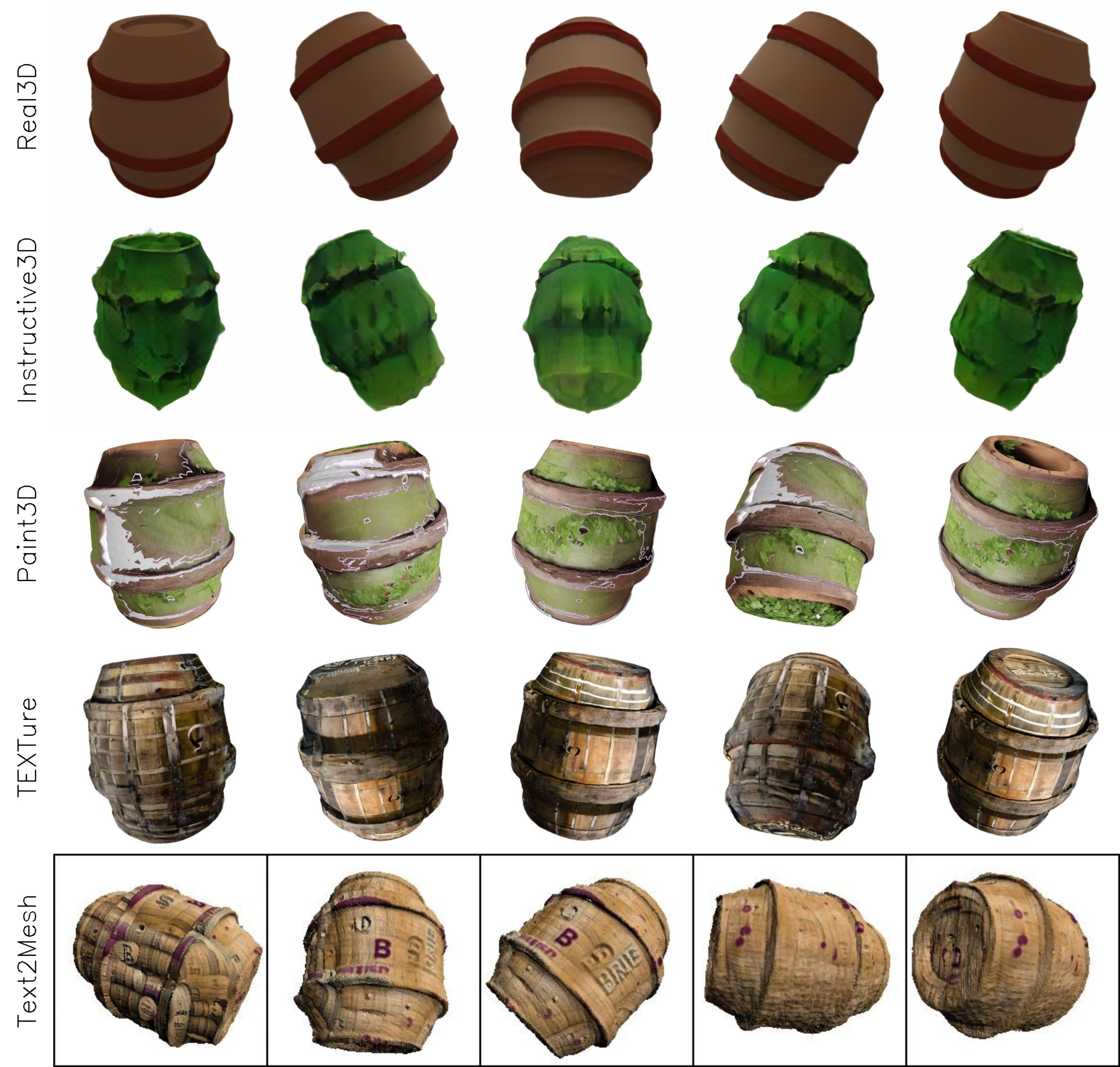}
	\caption{\textbf{Baseline comparison results.} Top row shows the rendered images from the mesh obtained from Real3D~\cite{jiang2024real3d}. Second row shows results from our method. Caption used for editing is: \textit{`apply leaves on the barrel''}.}
	\label{fig:supp_6}
\end{figure*}

\begin{figure*}
	\centering
	\includegraphics[width=\linewidth]{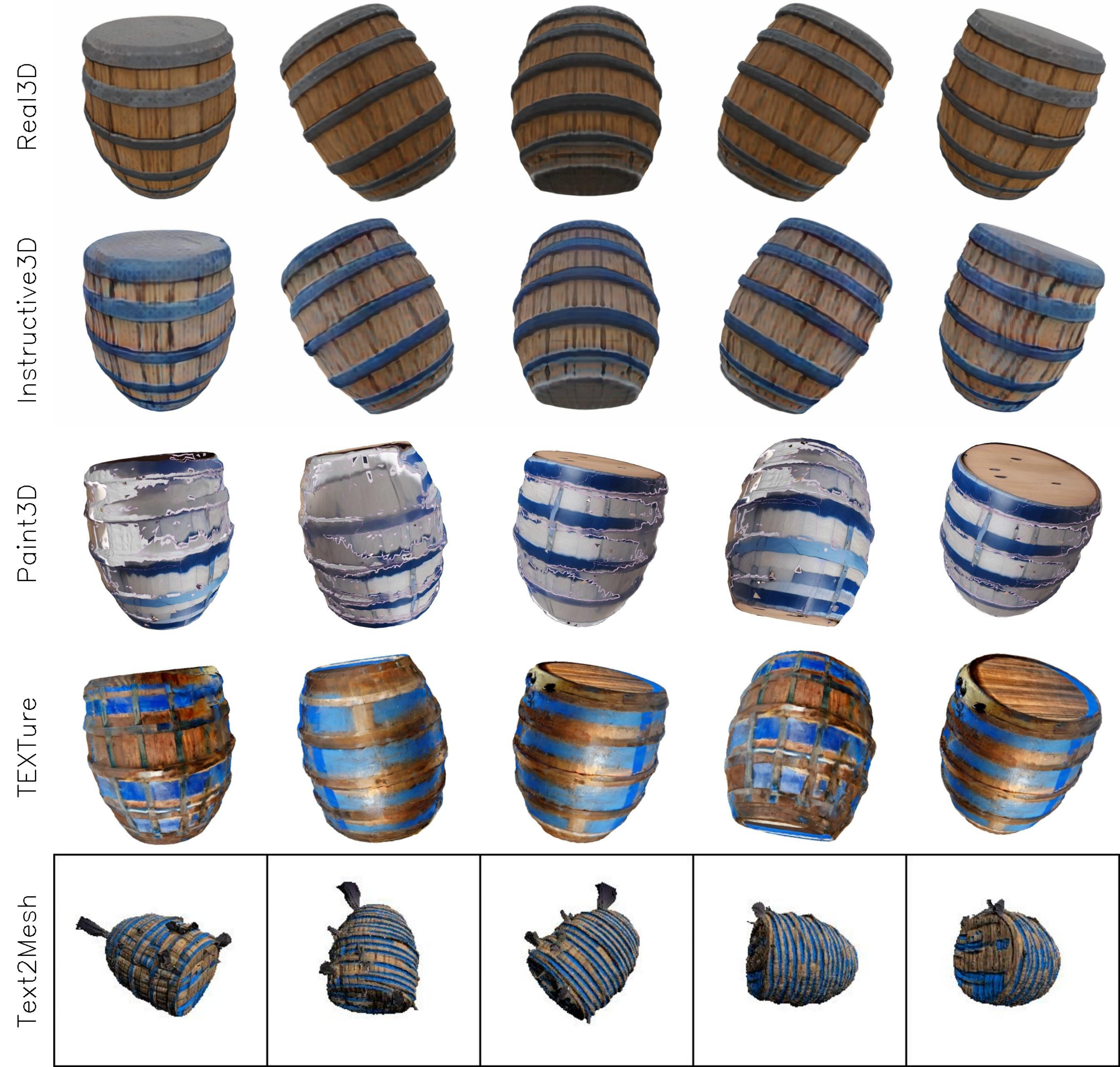}
	\caption{\textbf{Baseline comparison results.} Top row shows the rendered images from the mesh obtained from Real3D~\cite{jiang2024real3d}. Second row shows results from our method. Caption used for editing is: \textit{`add blue stripes to the barrel''}.}
	\label{fig:supp_7}
\end{figure*}

\begin{figure*}
	\centering
	\includegraphics[width=\linewidth]{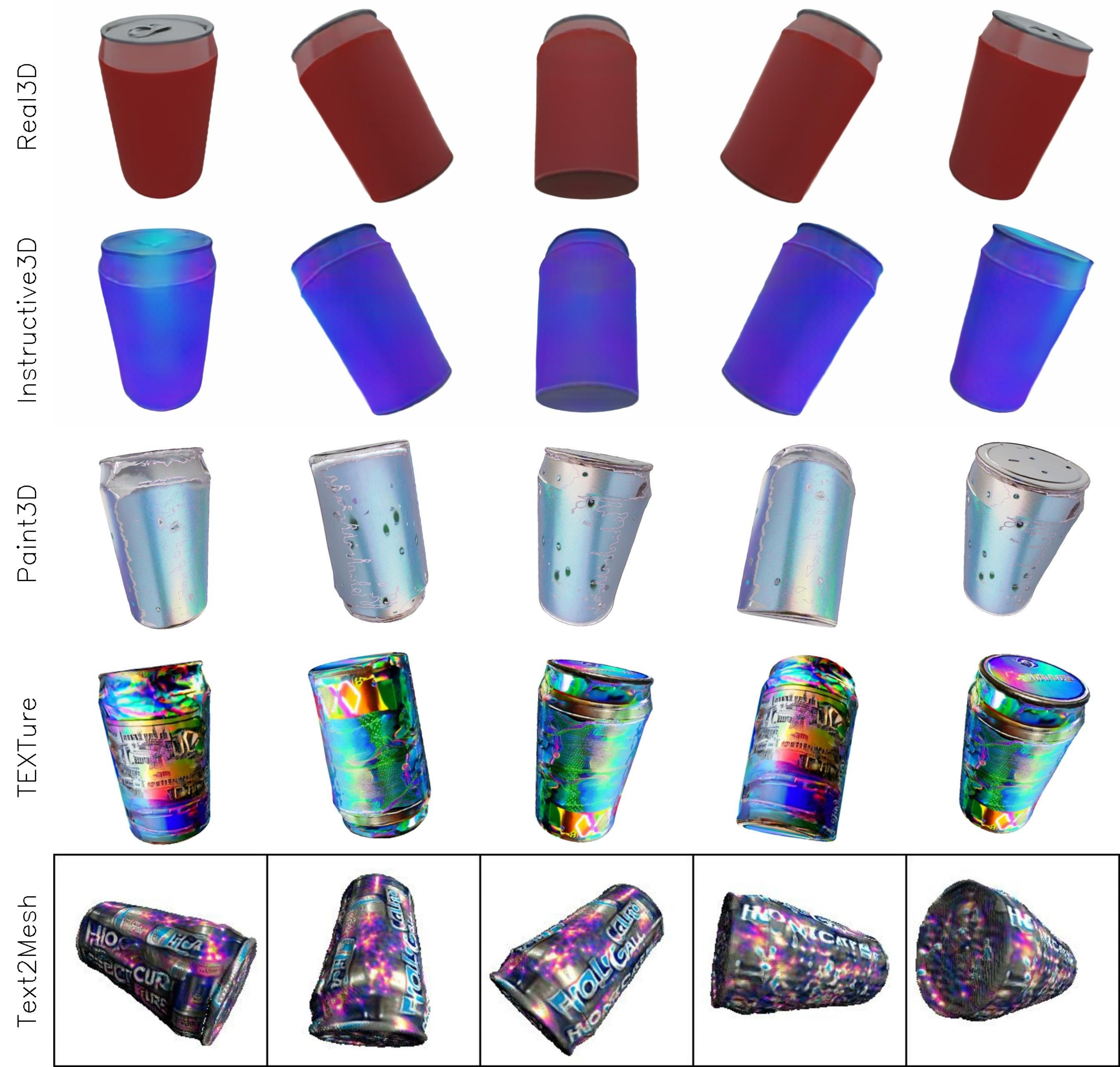}
	\caption{\textbf{Baseline comparison results.} Top row shows the rendered images from the mesh obtained from Real3D~\cite{jiang2024real3d}. Second row shows results from our method. Caption used for editing is: \textit{`apply a purple gradient color to can''}.}
	\label{fig:supp_8}
\end{figure*}

\begin{figure*}
	\centering
	\includegraphics[width=\linewidth]{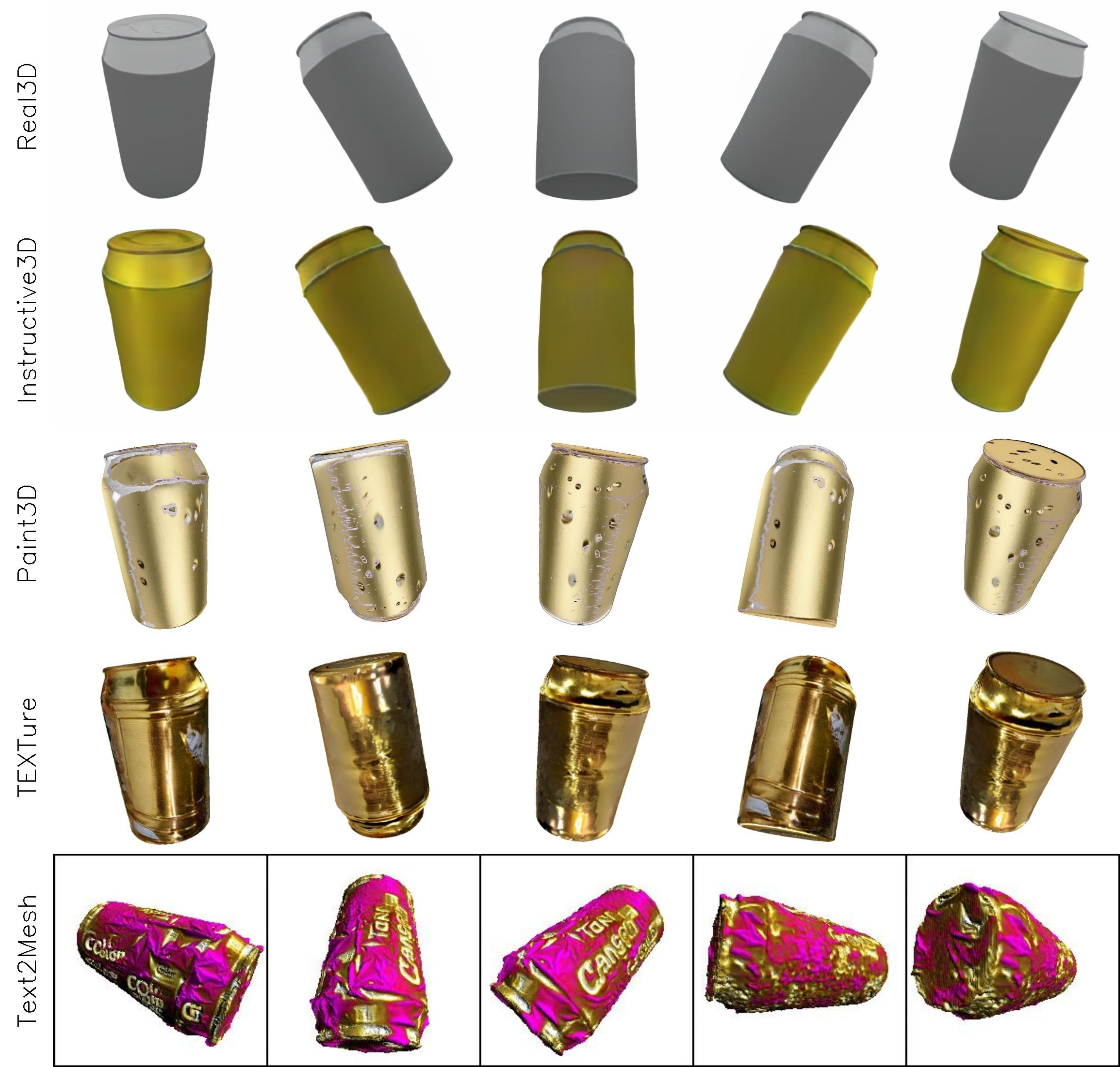}
	\caption{\textbf{Baseline comparison results.} Top row shows the rendered images from the mesh obtained from Real3D~\cite{jiang2024real3d}. Second row shows results from our method. Caption used for editing is: \textit{`change color of can to gold''}.}
	\label{fig:supp_9}
\end{figure*}

\begin{figure*}
	\centering
	\includegraphics[width=\linewidth]{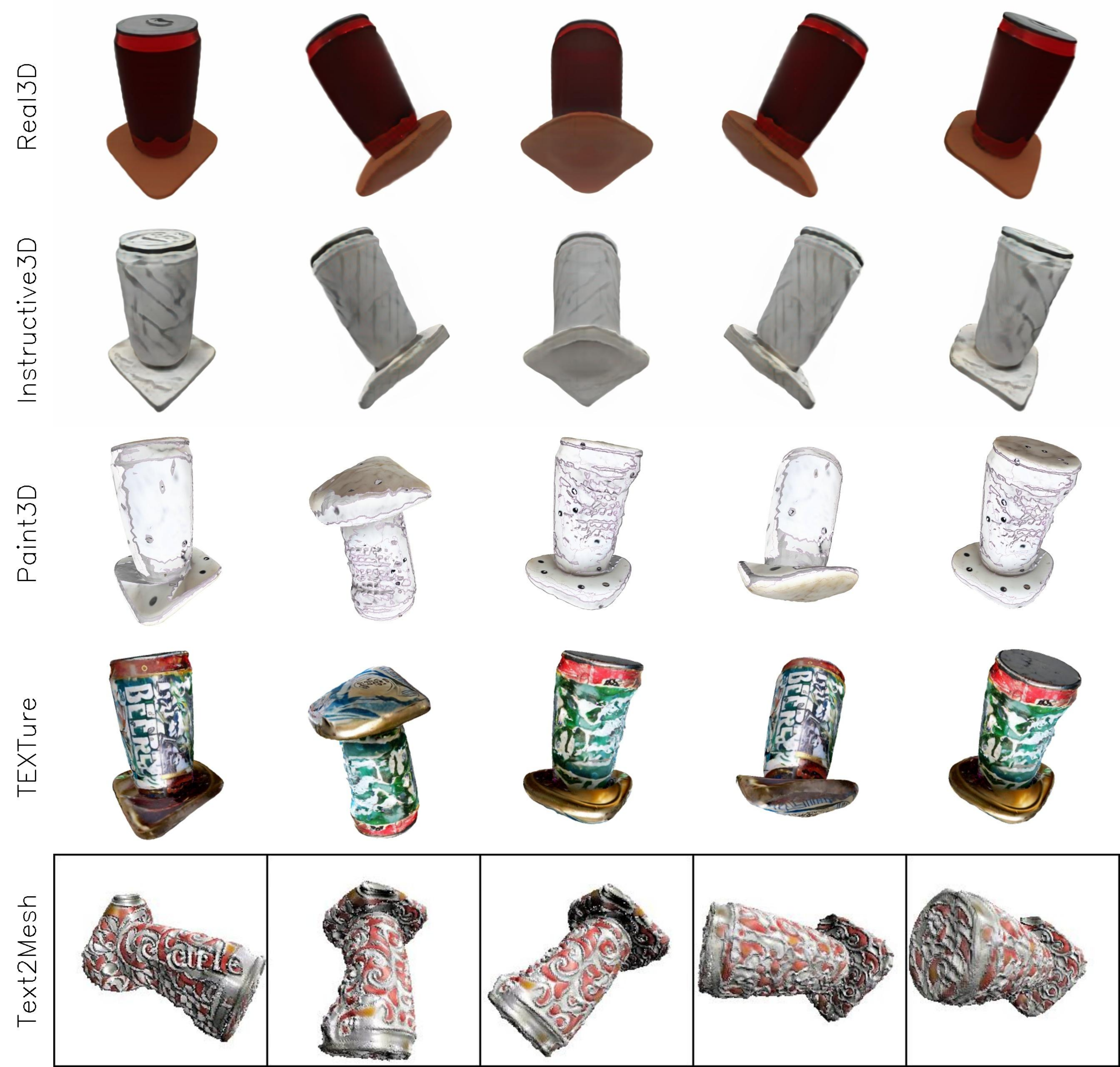}
	\caption{\textbf{Baseline comparison results.} Top row shows the rendered images from the mesh obtained from Real3D~\cite{jiang2024real3d}. Second row shows results from our method. Caption used for editing is: \textit{`add a marble effect to the can''}.}
	\label{fig:supp_10}
\end{figure*}

\begin{figure*}
	\centering
	\includegraphics[width=\linewidth]{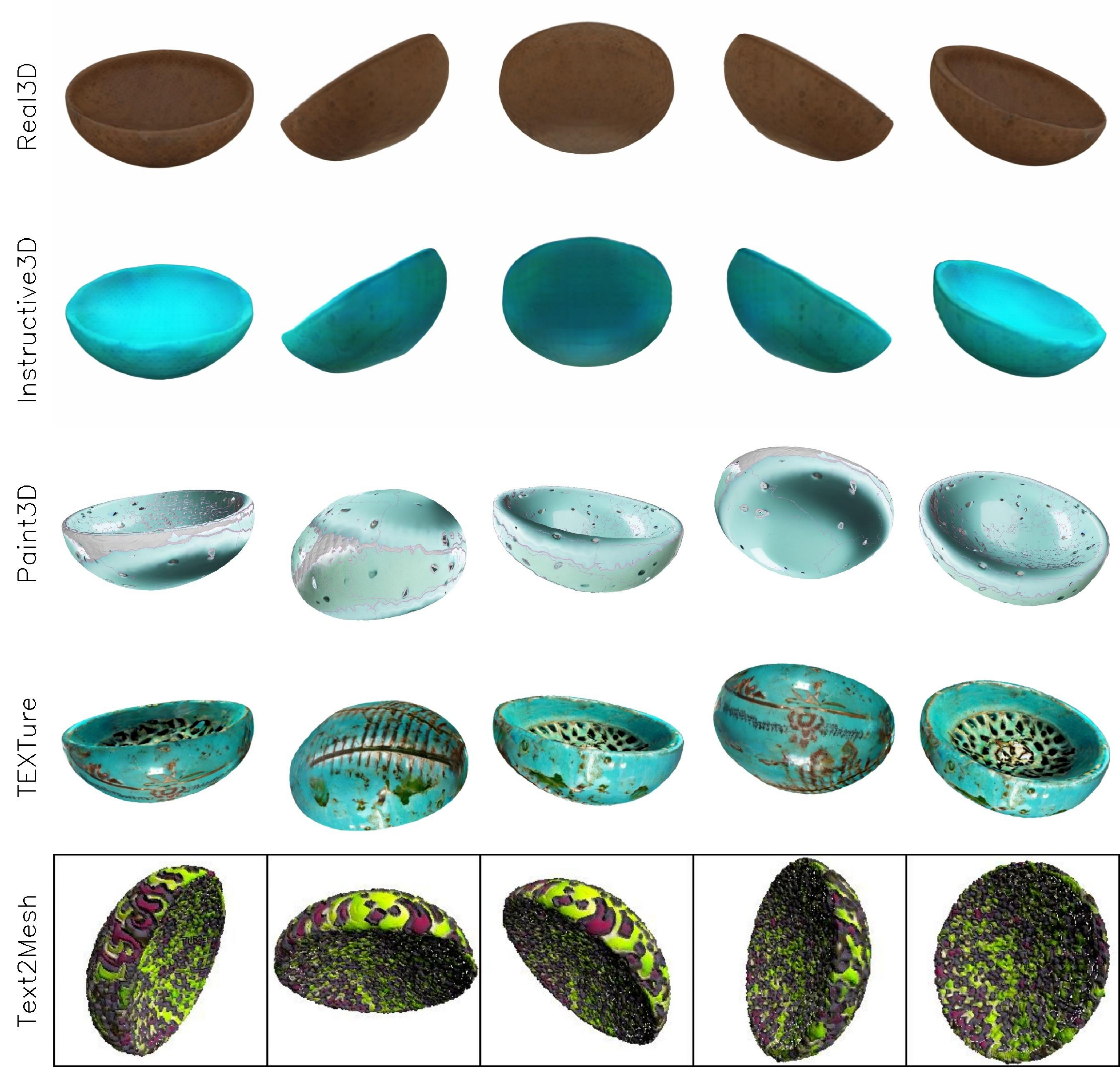}
	\caption{\textbf{Baseline comparison results.} Top row shows the rendered images from the mesh obtained from Real3D~\cite{jiang2024real3d}. Second row shows results from our method. Caption used for editing is: \textit{`change color of bowl to turqoise''}.}
	\label{fig:supp_11}
\end{figure*}

\begin{figure*}
	\centering
	\includegraphics[width=\linewidth]{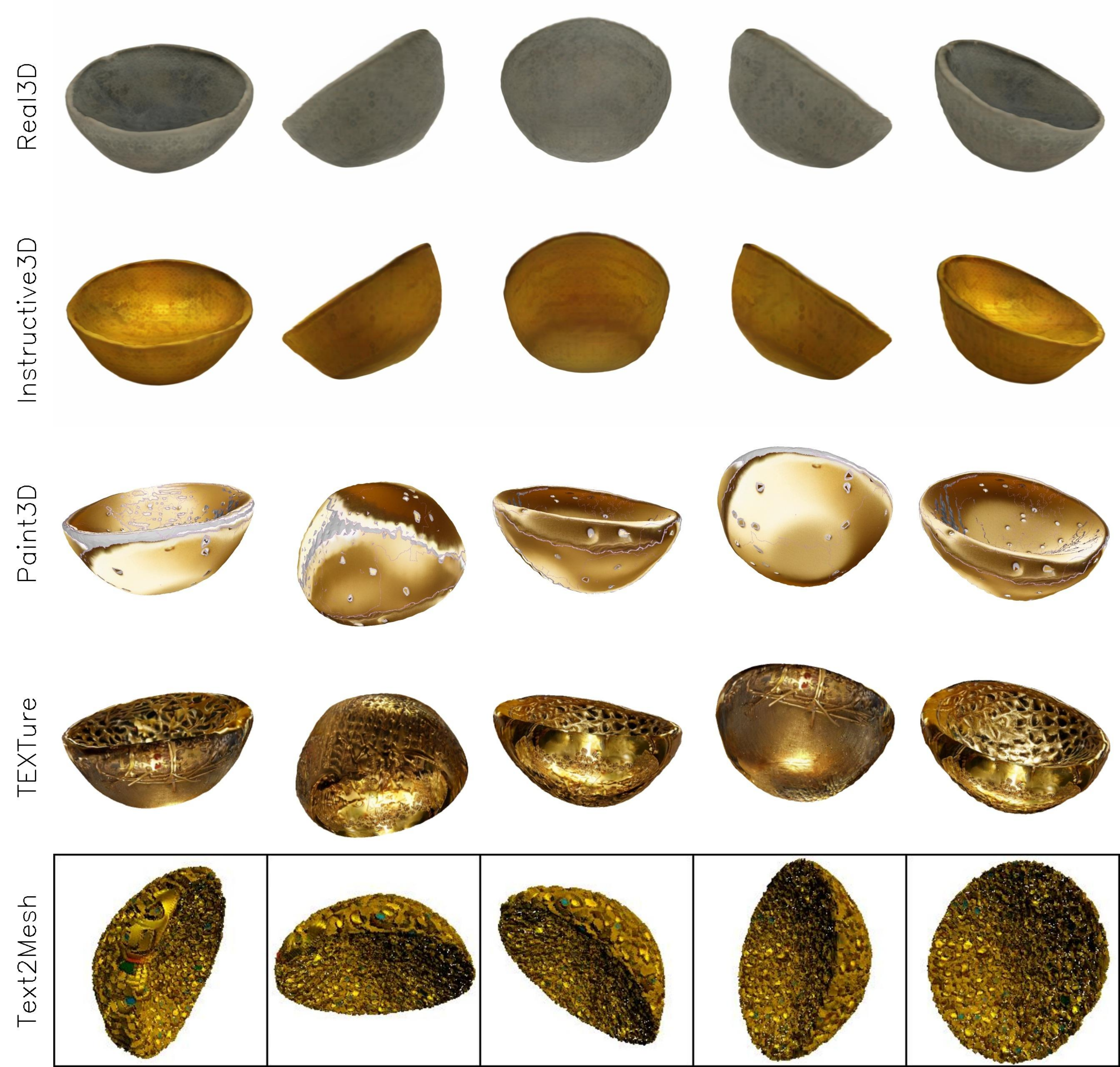}
	\caption{\textbf{Baseline comparison results.} Top row shows the rendered images from the mesh obtained from Real3D~\cite{jiang2024real3d}. Second row shows results from our method. Caption used for editing is: \textit{`change color of bowl to gold''}.}
	\label{fig:supp_12}
\end{figure*}

\begin{figure*}
	\centering
	\includegraphics[width=\linewidth]{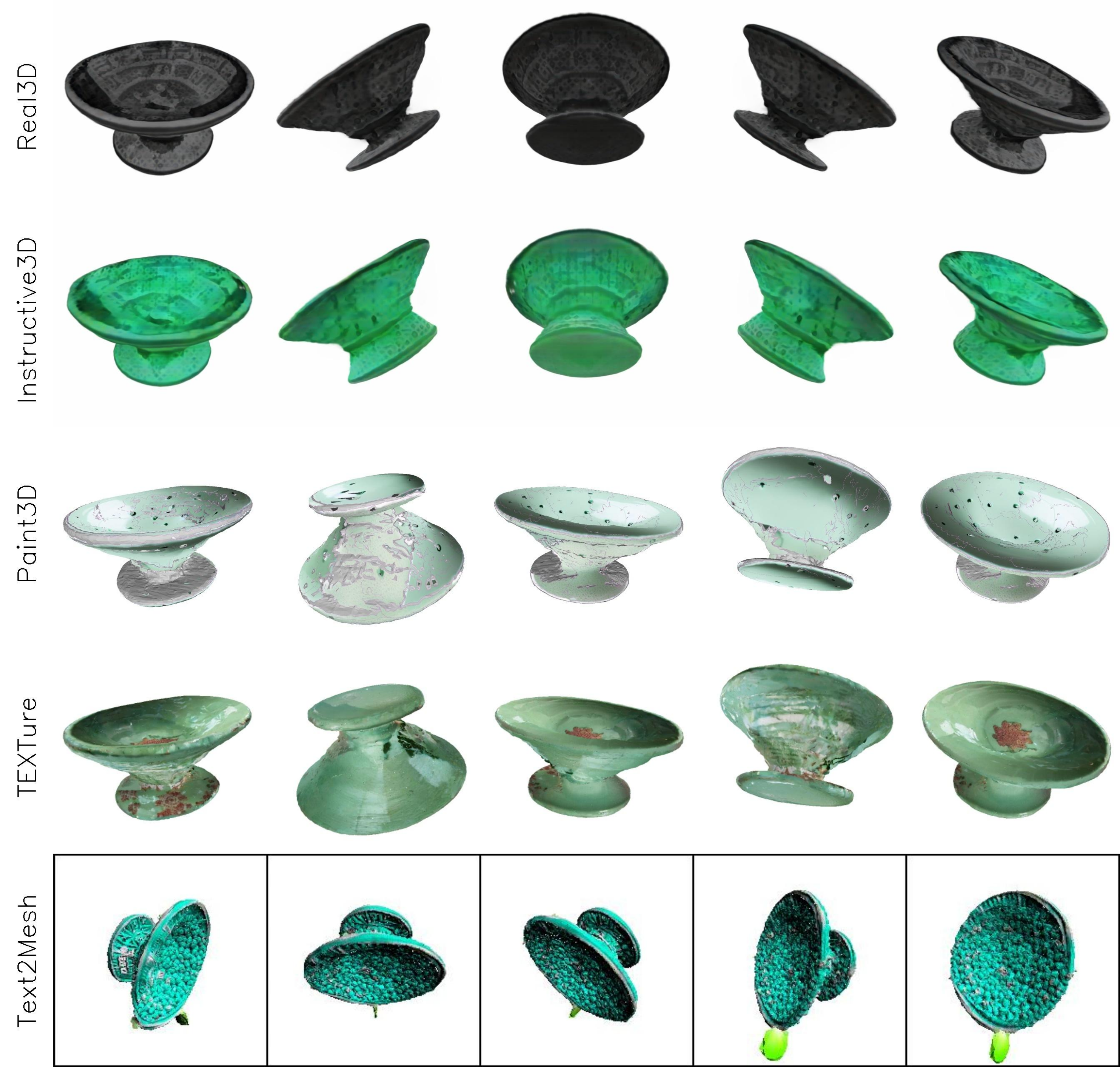}
	\caption{\textbf{Baseline comparison results.} Top row shows the rendered images from the mesh obtained from Real3D~\cite{jiang2024real3d}. Second row shows results from our method. Caption used for editing is: \textit{`change color of bowl to mint green''}.}
	\label{fig:supp_13}
\end{figure*}

\begin{figure*}
	\centering
	\includegraphics[width=\linewidth]{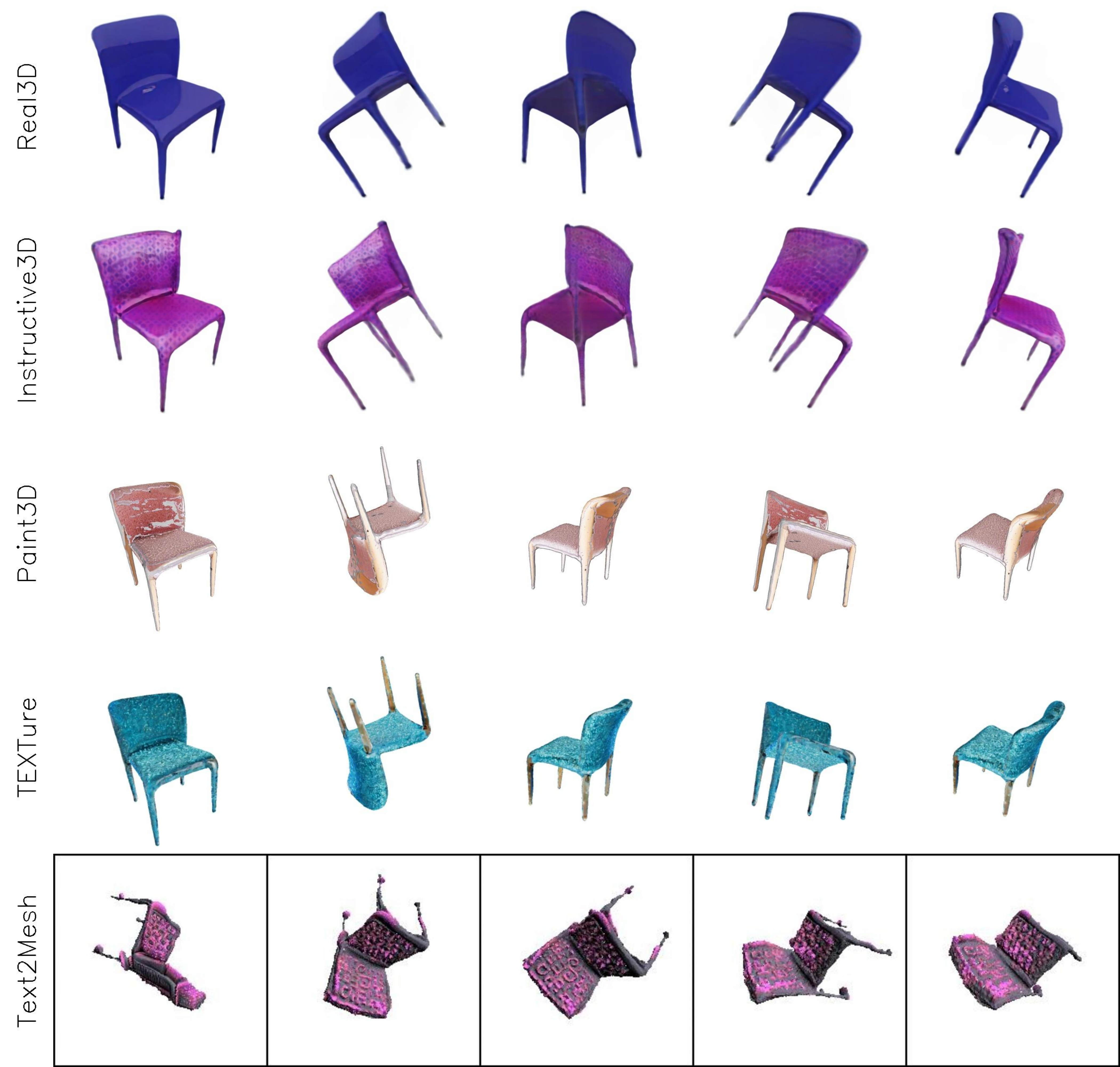}
	\caption{\textbf{Baseline comparison results.} Top row shows the rendered images from the mesh obtained from Real3D~\cite{jiang2024real3d}. Second row shows results from our method. Caption used for editing is: \textit{``add a purple glittery look to chair''}.}
	\label{fig:supp_14}
\end{figure*}

\begin{figure*}
	\centering
	\includegraphics[width=\linewidth]{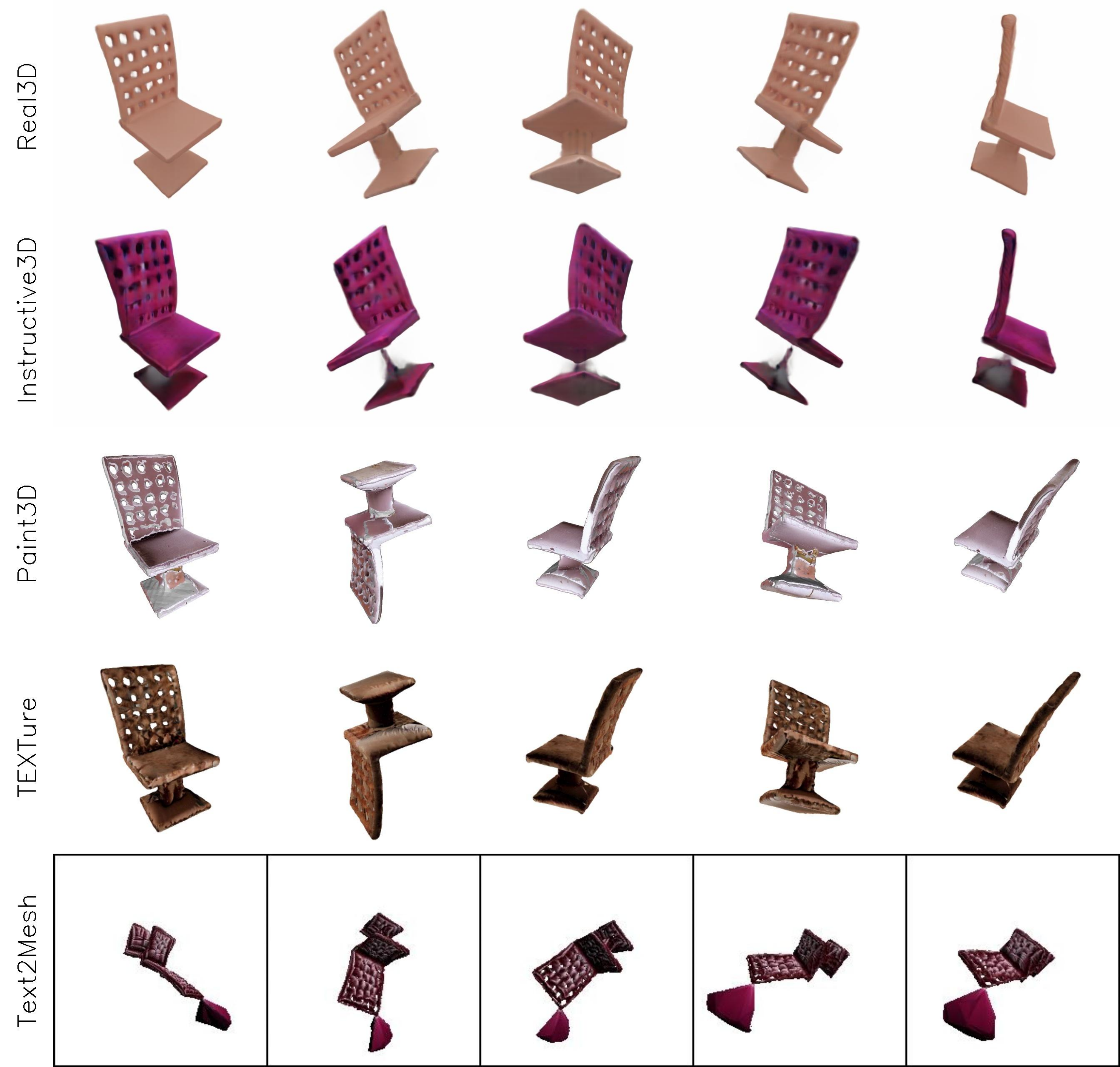}
	\caption{\textbf{Baseline comparison results.} Top row shows the rendered images from the mesh obtained from Real3D~\cite{jiang2024real3d}. Second row shows results from our method. Caption used for editing is: \textit{``add a velvet texture to the chair''}.}
	\label{fig:supp_15}
\end{figure*}

\begin{figure*}
	\centering
	\includegraphics[width=\linewidth]{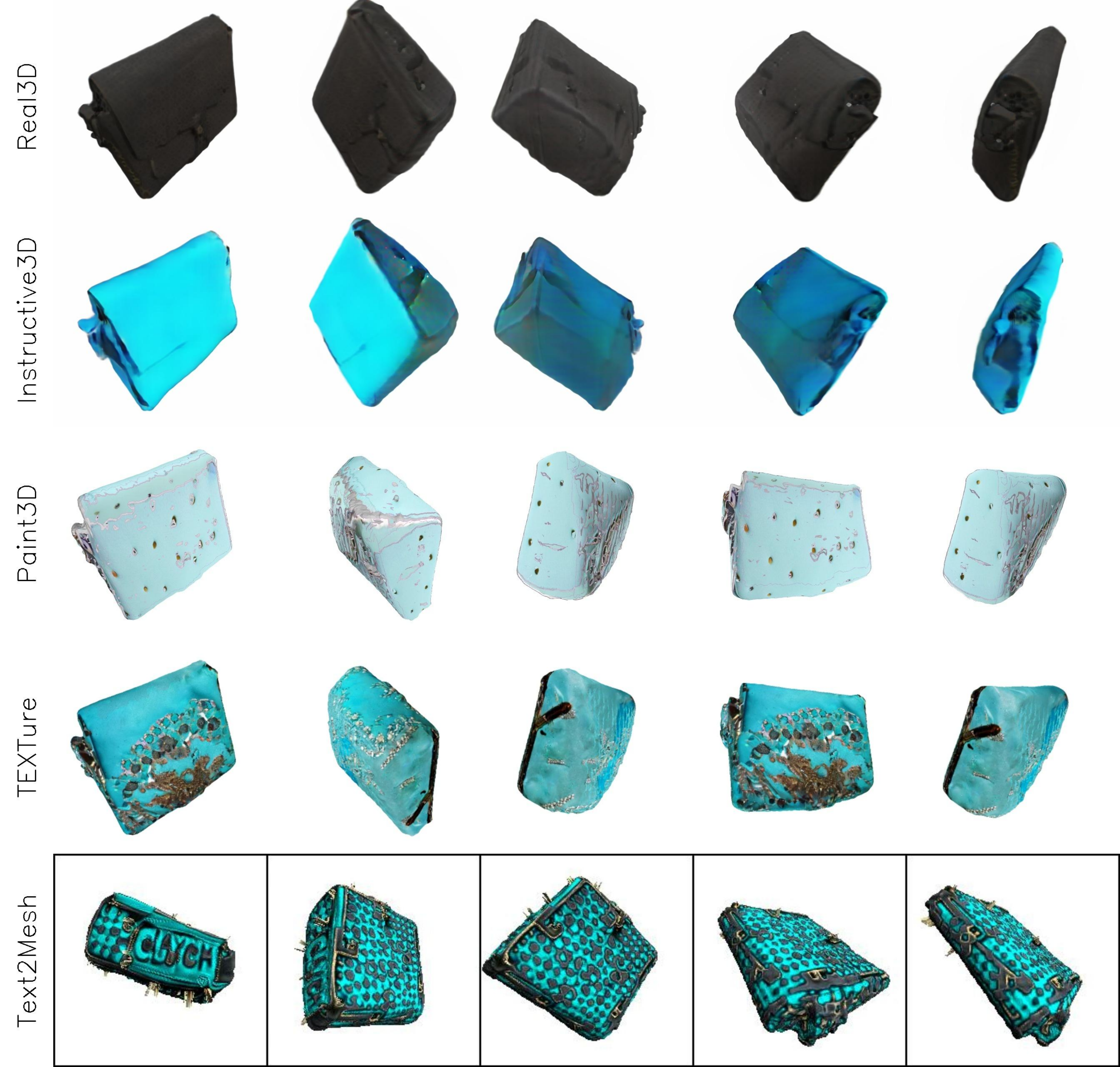}
	\caption{\textbf{Baseline comparison results.} Top row shows the rendered images from the mesh obtained from Real3D~\cite{jiang2024real3d}. Second row shows results from our method. Caption used for editing is: \textit{``change color of clutch bag to cyan''}.}
	\label{fig:supp_16}
\end{figure*}

\begin{figure*}
	\centering
	\includegraphics[width=\linewidth]{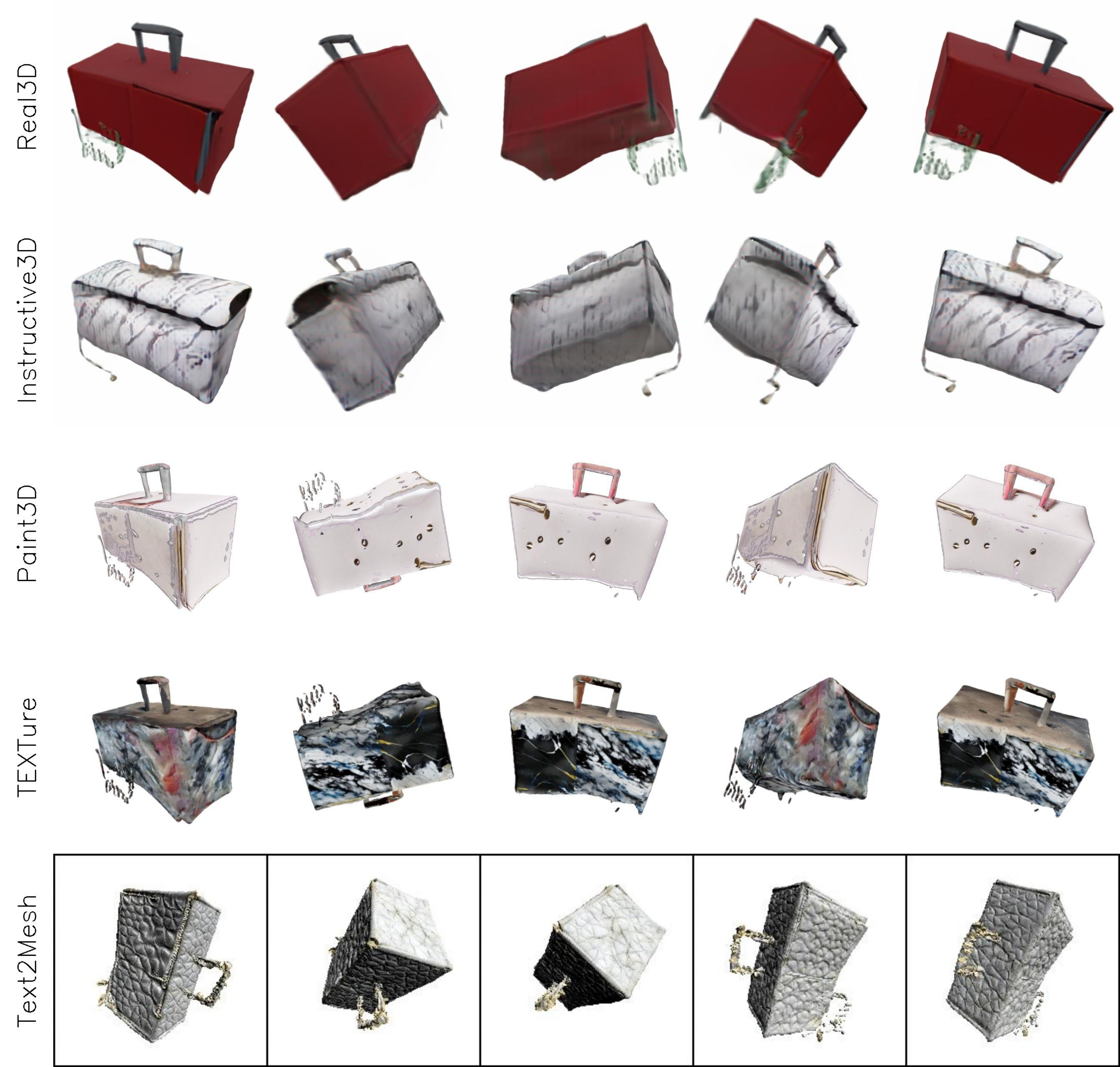}
	\caption{\textbf{Baseline comparison results.} Top row shows the rendered images from the mesh obtained from Real3D~\cite{jiang2024real3d}. Second row shows results from our method. Caption used for editing is: \textit{``apply marble texture to the clutch bag''}.}
	\label{fig:supp_17}
\end{figure*}

\begin{figure*}
	\centering
	\includegraphics[width=\linewidth]{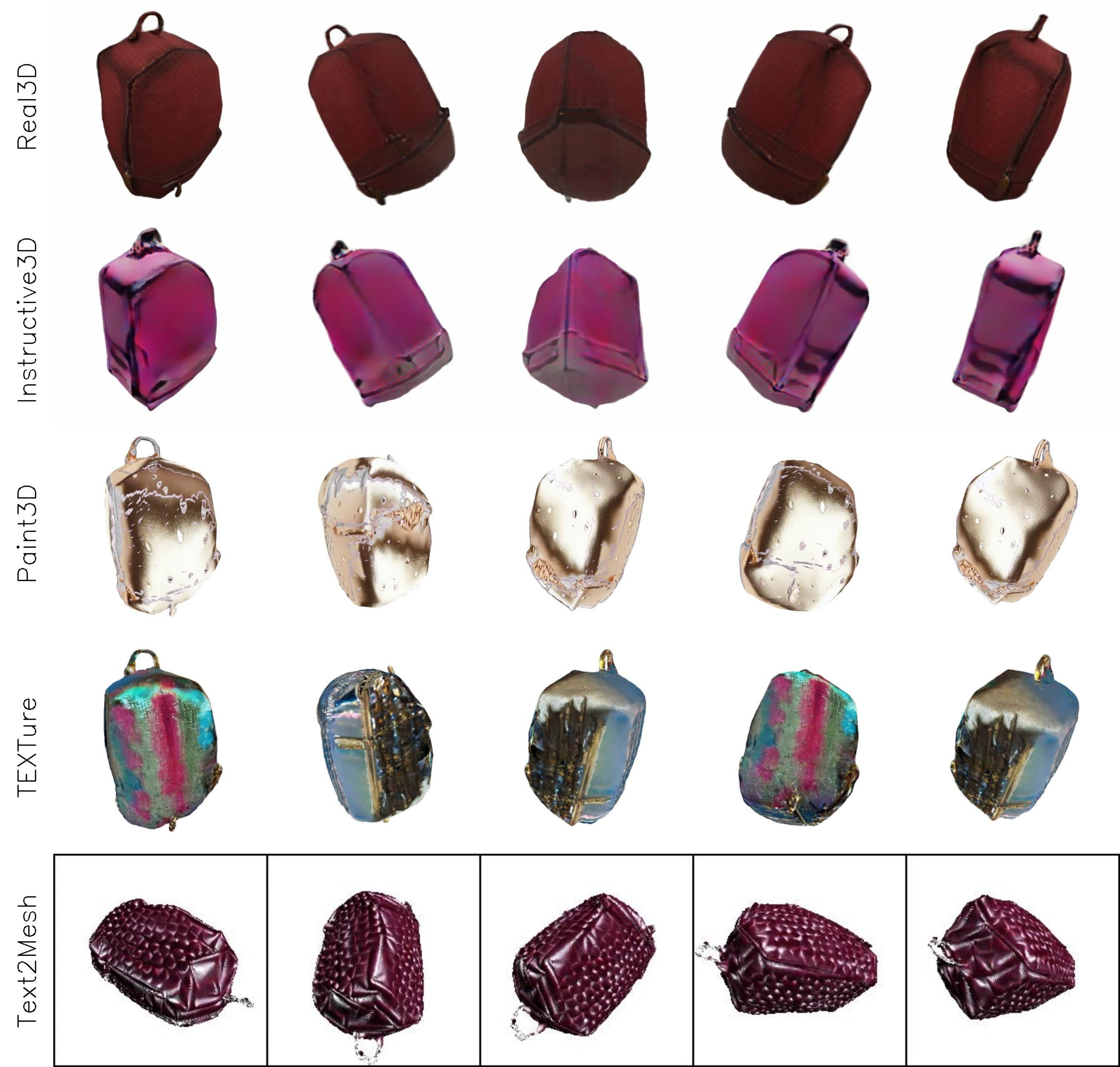}
	\caption{\textbf{Baseline comparison results.} Top row shows the rendered images from the mesh obtained from Real3D~\cite{jiang2024real3d}. Second row shows results from our method. Caption used for editing is: \textit{``add a glossy texture to the clutch bag''}.}
	\label{fig:supp_18}
\end{figure*}

\begin{figure*}
	\centering
	\includegraphics[width=\linewidth]{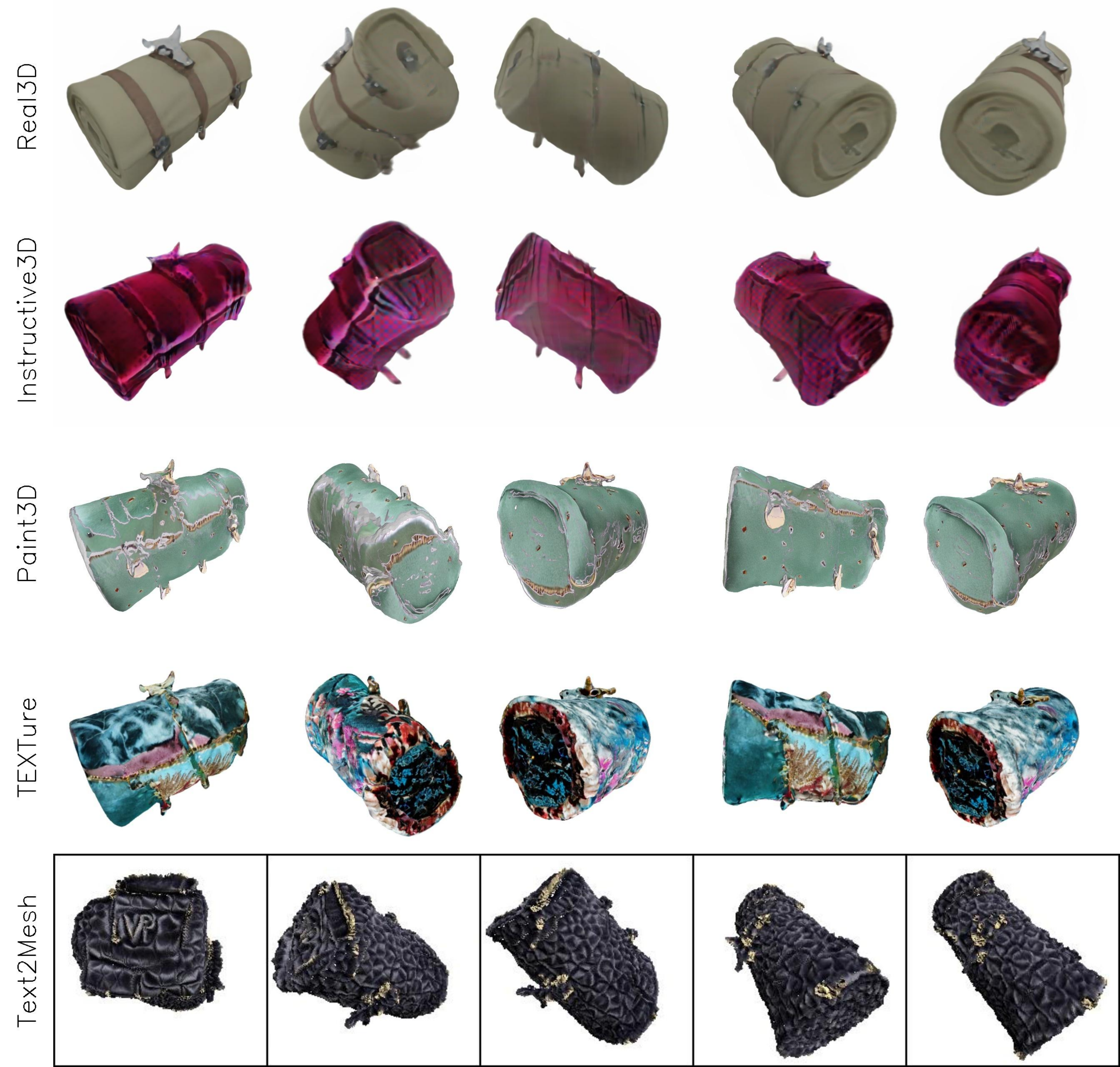}
	\caption{\textbf{Baseline comparison results.} Top row shows the rendered images from the mesh obtained from Real3D~\cite{jiang2024real3d}. Second row shows results from our method. Caption used for editing is: \textit{``add a velvet texture to the clutch bag''}.}
	\label{fig:supp_19}
\end{figure*}

\begin{figure*}
	\centering
	\includegraphics[width=\linewidth]{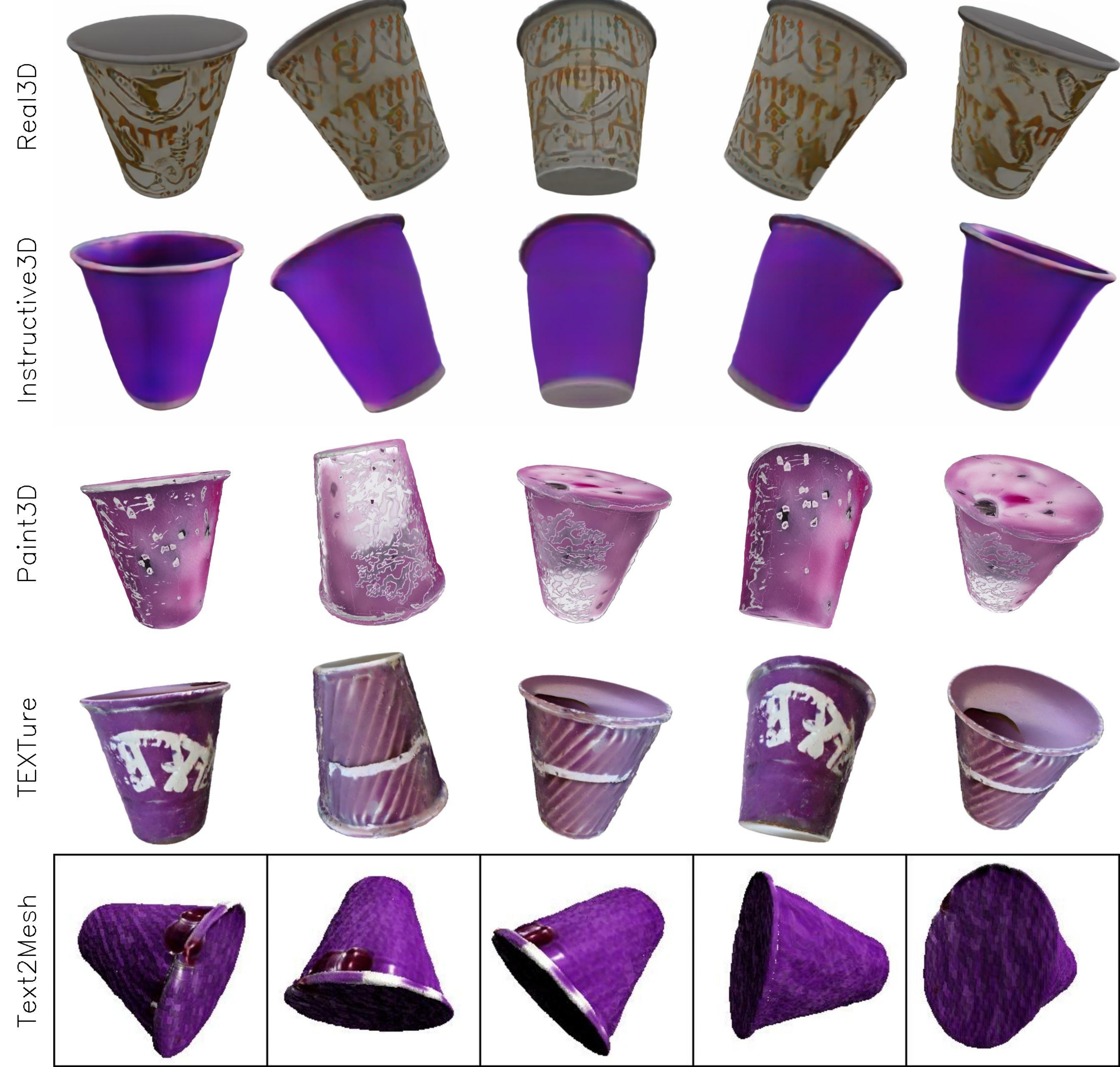}
	\caption{\textbf{Baseline comparison results.} Top row shows the rendered images from the mesh obtained from Real3D~\cite{jiang2024real3d}. Second row shows results from our method. Caption used for editing is: \textit{``change color to purple''}.}
	\label{fig:supp_20}
\end{figure*}

\begin{figure*}
	\centering
	\includegraphics[width=\linewidth]{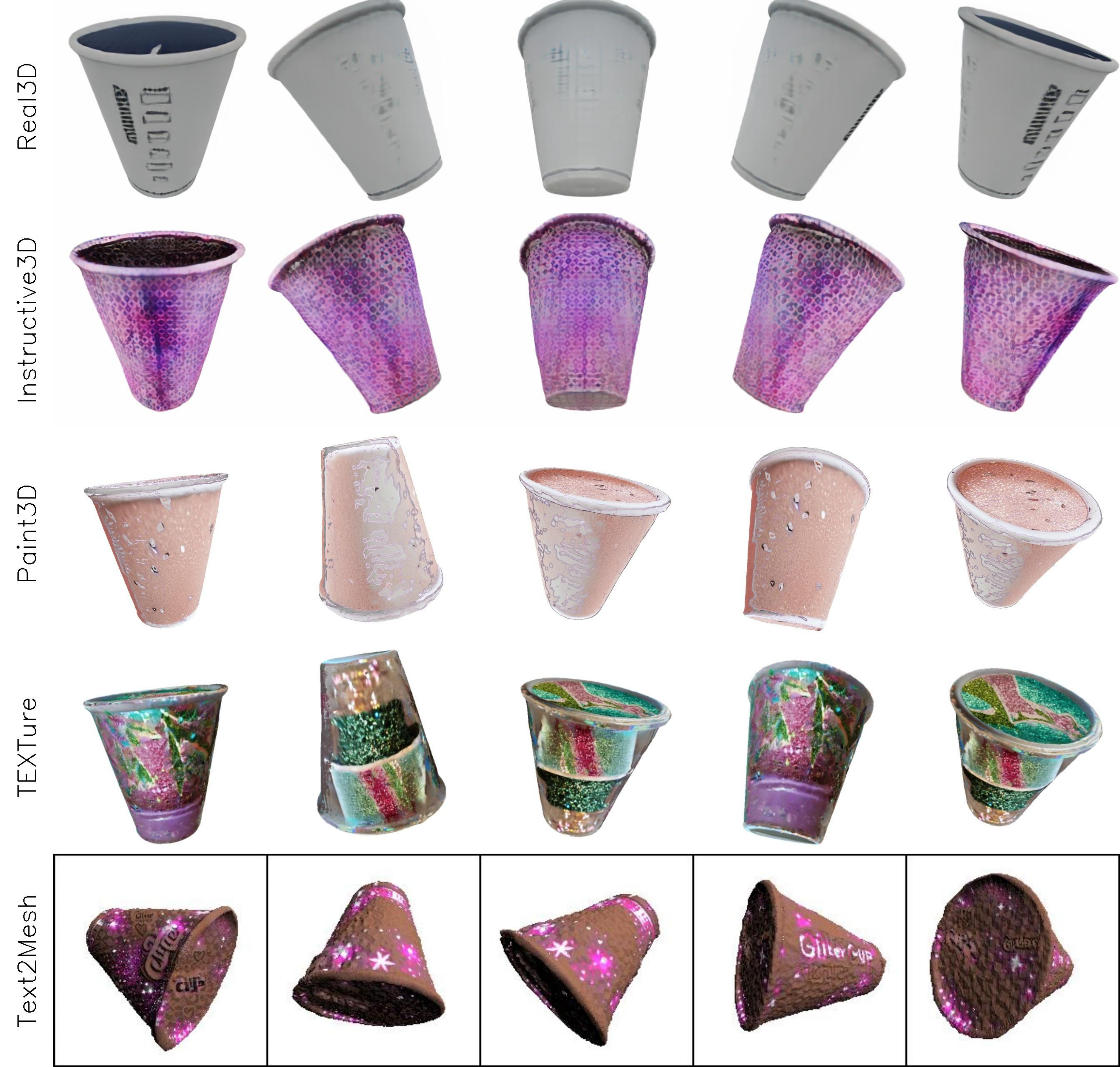}
	\caption{\textbf{Baseline comparison results.} Top row shows the rendered images from the mesh obtained from Real3D~\cite{jiang2024real3d}. Second row shows results from our method. Caption used for editing is: \textit{``add a glittery pink overlay to the cup''}.}
	\label{fig:supp_21}
\end{figure*}

\begin{figure*}
	\centering
	\includegraphics[width=\linewidth]{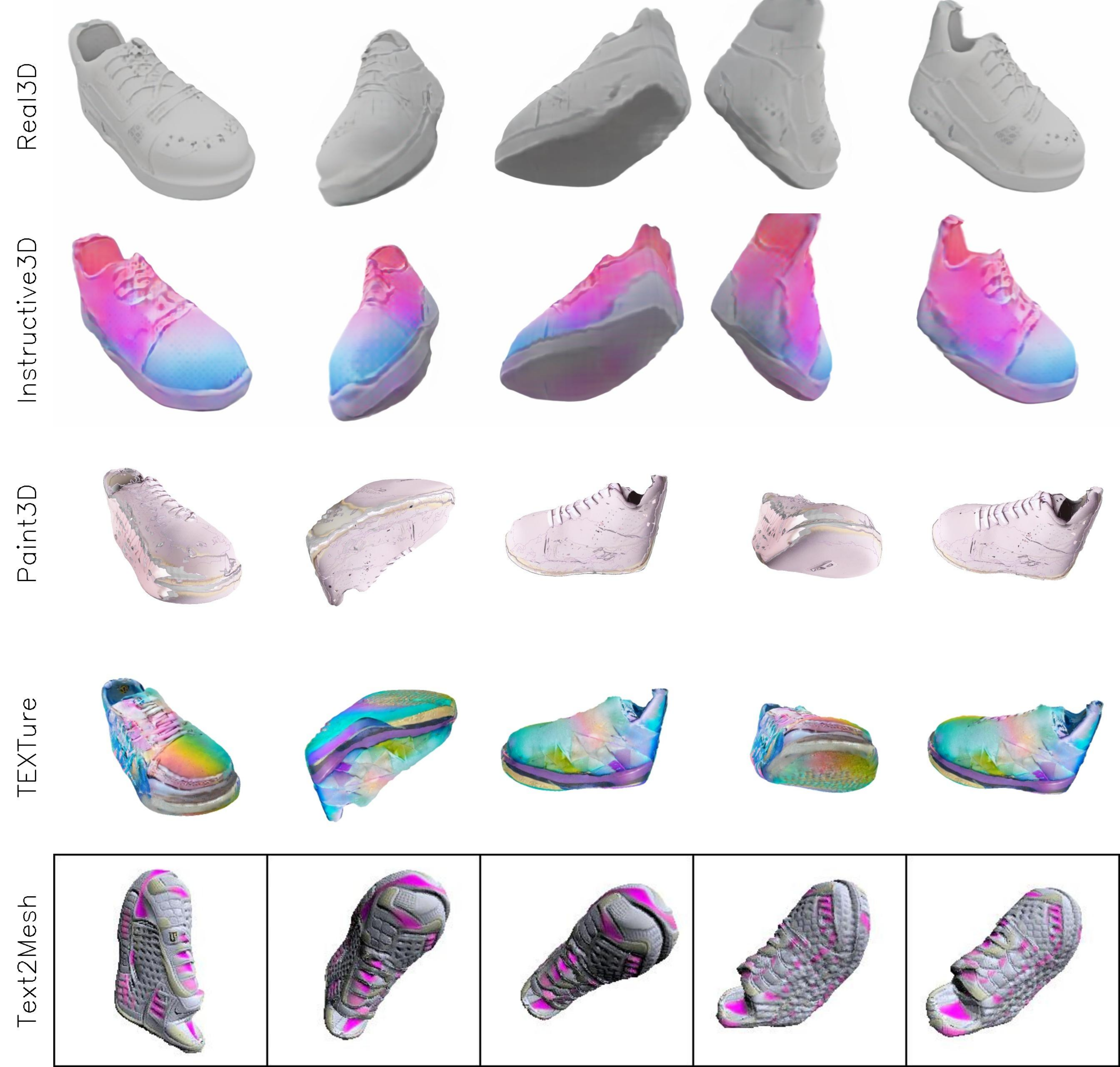}
	\caption{\textbf{Baseline comparison results.} Top row shows the rendered images from the mesh obtained from Real3D~\cite{jiang2024real3d}. Second row shows results from our method. Caption used for editing is: \textit{``add a pastel gradient to the shoe''}.}
	\label{fig:supp_22}
\end{figure*}

\begin{figure*}
	\centering
	\includegraphics[width=\linewidth]{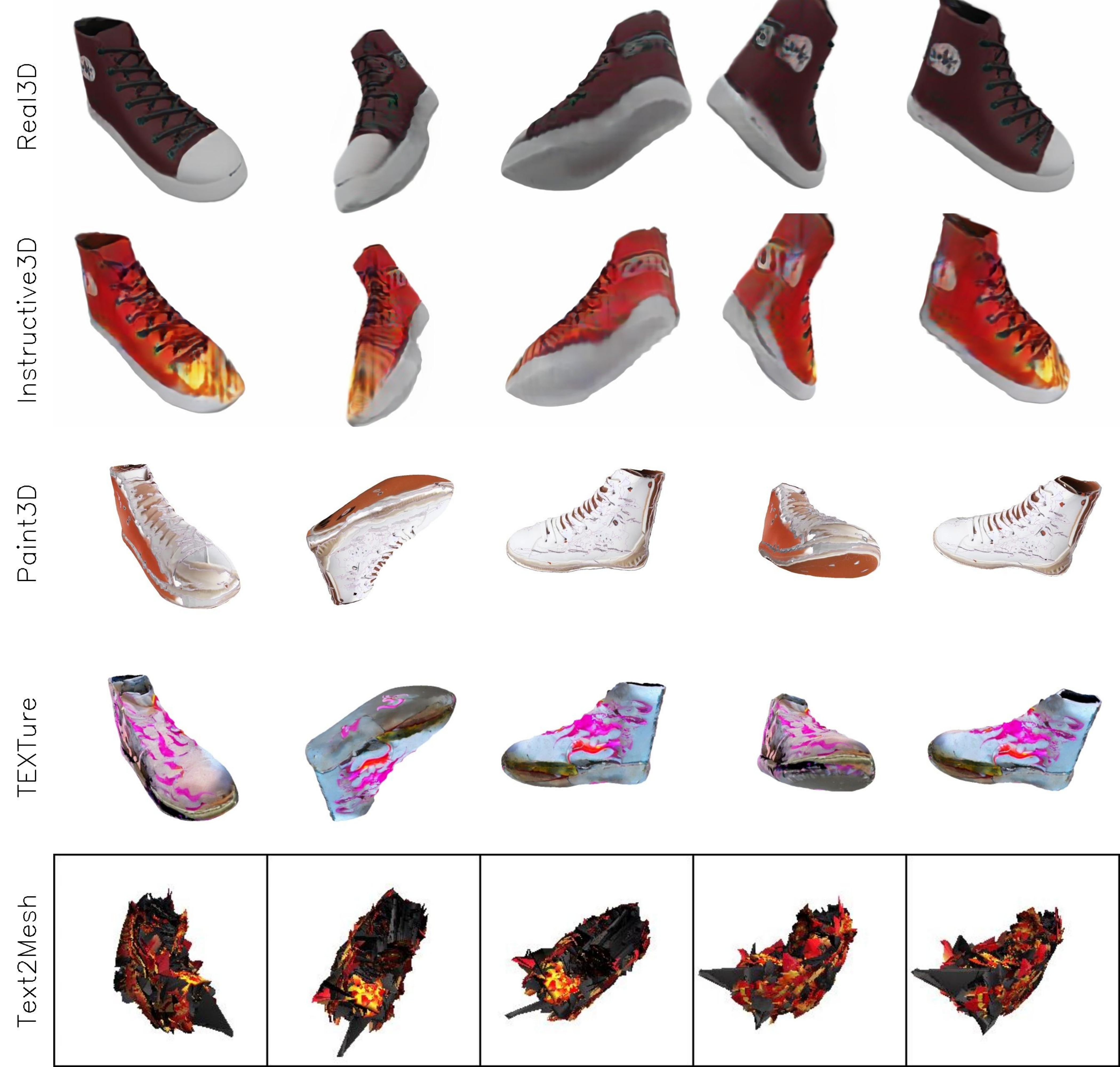}
	\caption{\textbf{Baseline comparison results.} Top row shows the rendered images from the mesh obtained from Real3D~\cite{jiang2024real3d}. Second row shows results from our method. Caption used for editing is: \textit{``add a flame design to the shoe''}.}
	\label{fig:supp_23}
\end{figure*}

\begin{figure*}
	\centering
	\includegraphics[width=\linewidth]{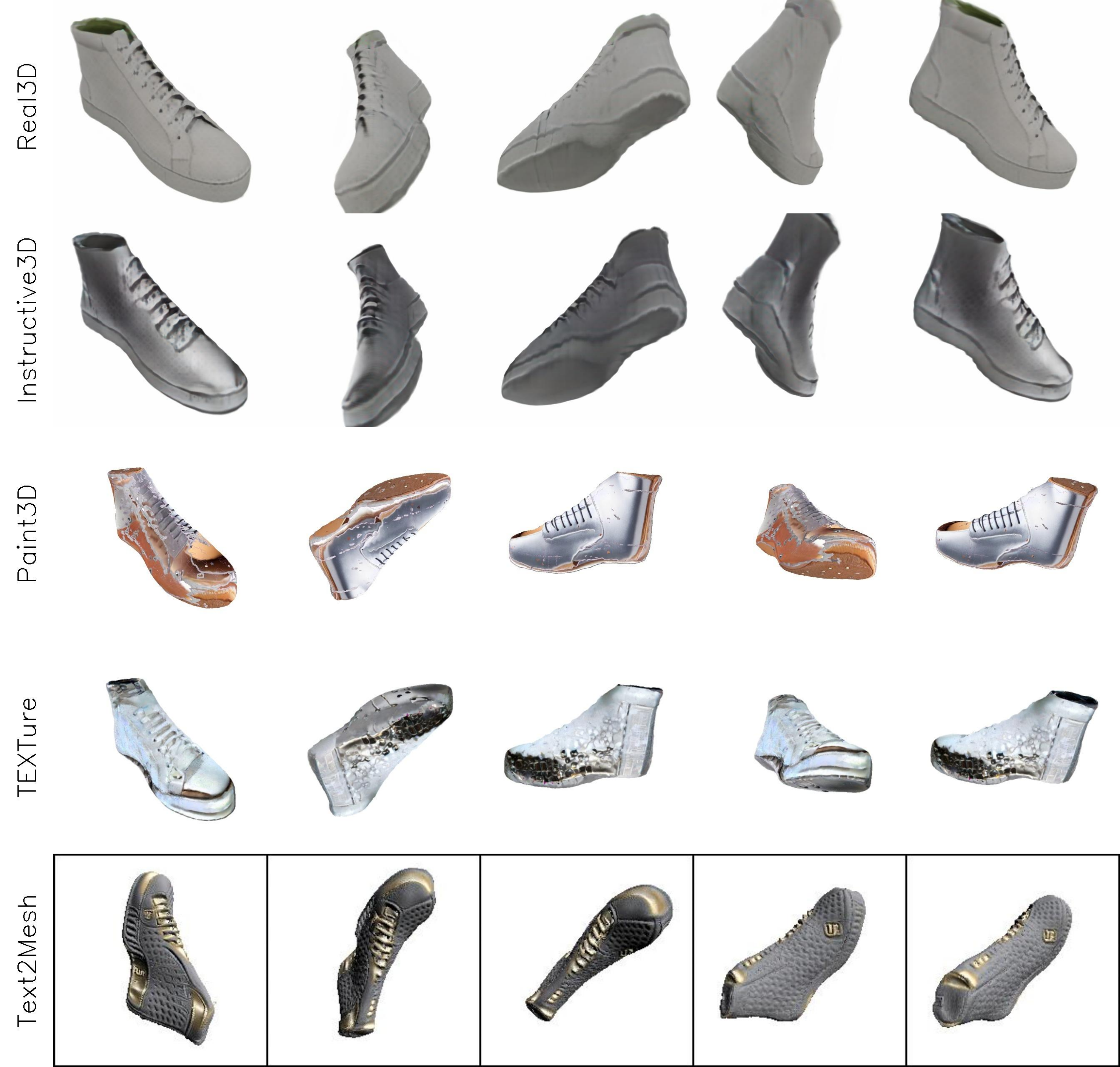}
	\caption{\textbf{Baseline comparison results.} Top row shows the rendered images from the mesh obtained from Real3D~\cite{jiang2024real3d}. Second row shows results from our method. Caption used for editing is: \textit{``add a brushed metal finish to the shoe''}.}
	\label{fig:supp_24}
\end{figure*}

\begin{figure*}
	\centering
	\includegraphics[width=\linewidth]{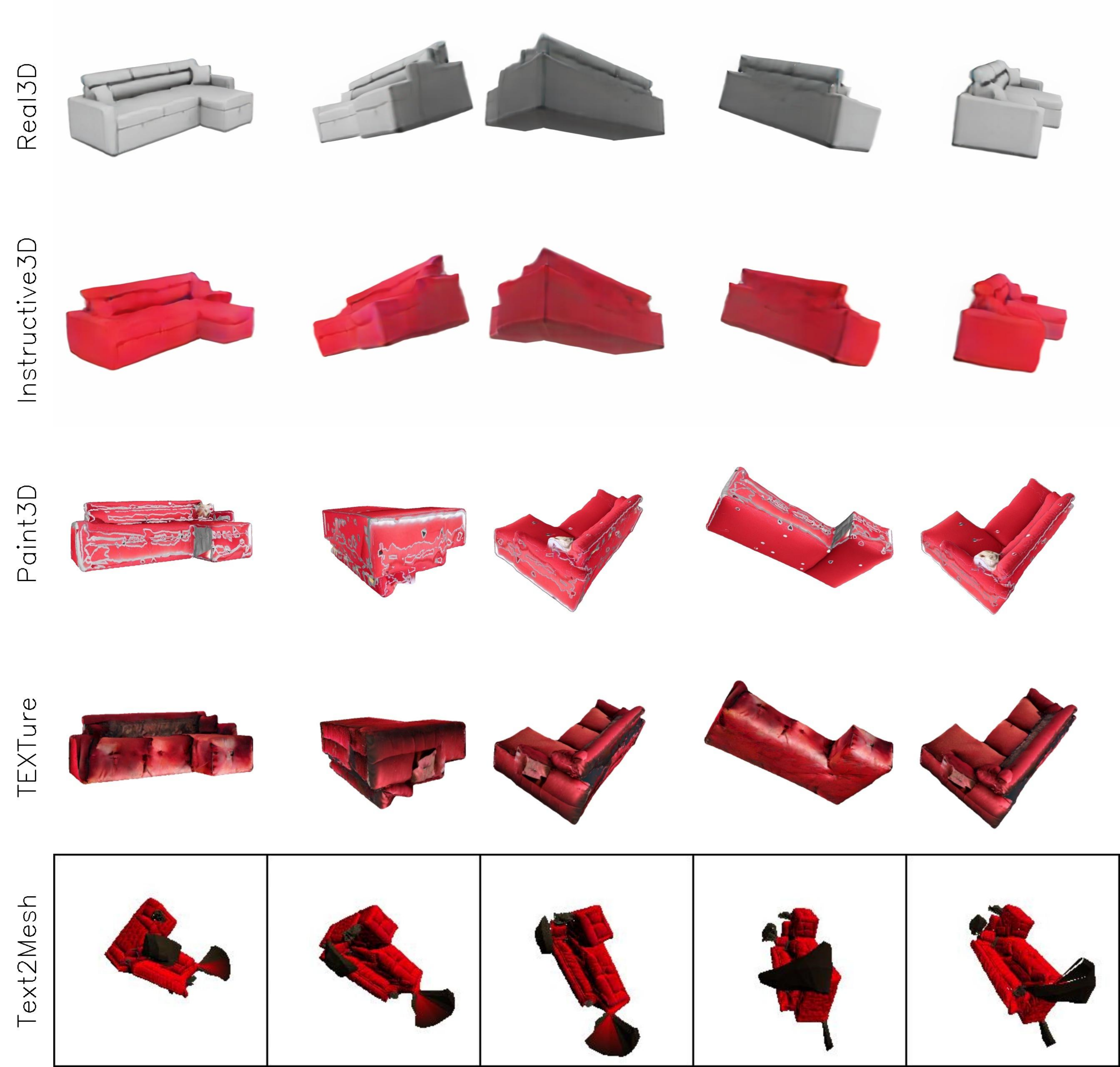}
	\caption{\textbf{Baseline comparison results.} Top row shows the rendered images from the mesh obtained from Real3D~\cite{jiang2024real3d}. Second row shows results from our method. Caption used for editing is: \textit{``change the color of sofa to red''}.}
	\label{fig:supp_25}
\end{figure*}

\begin{figure*}
	\centering
	\includegraphics[width=\linewidth]{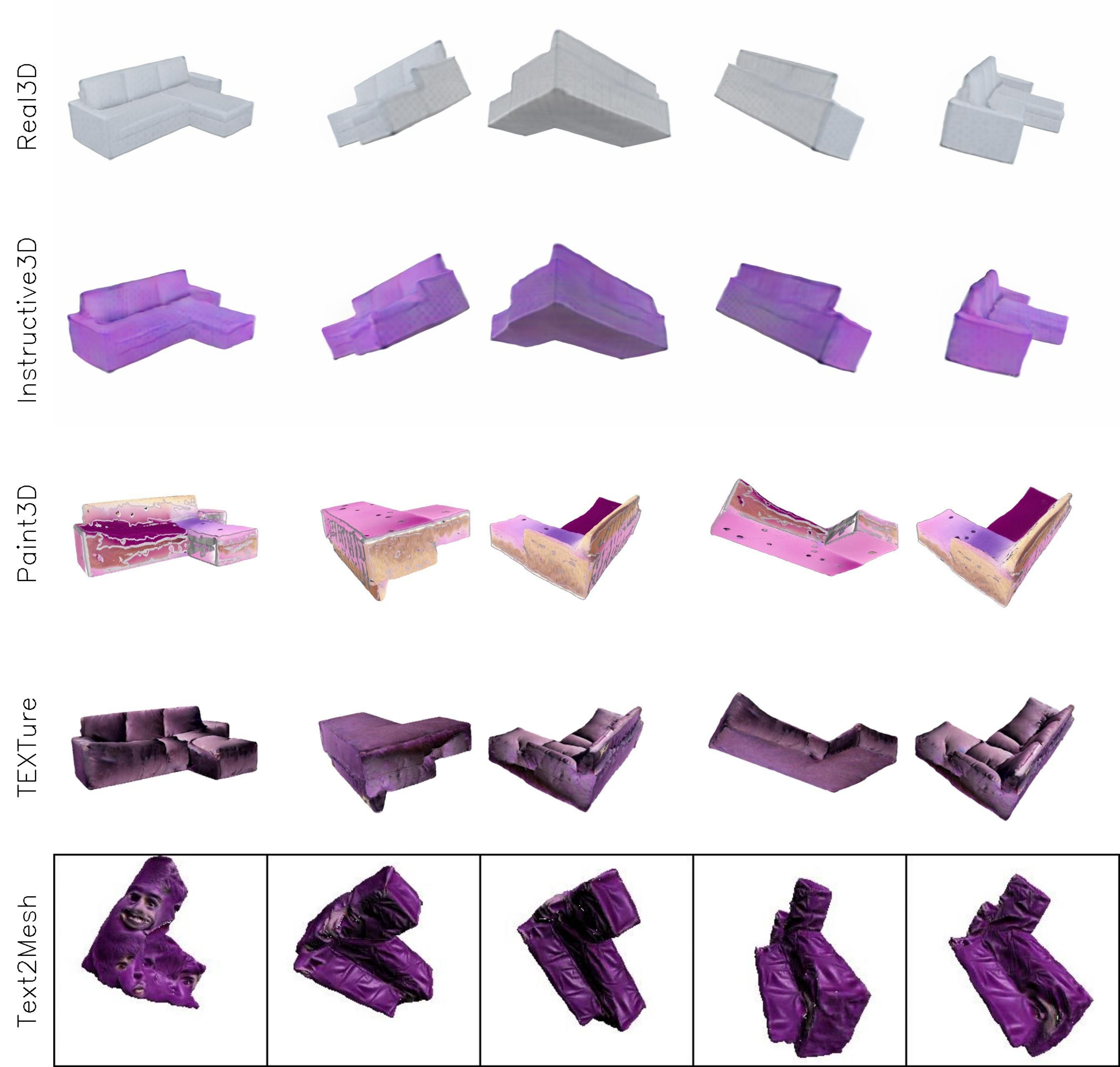}
	\caption{\textbf{Baseline comparison results.} Top row shows the rendered images from the mesh obtained from Real3D~\cite{jiang2024real3d}. Second row shows results from our method. Caption used for editing is: \textit{``change color of sofa to purple''}.}
	\label{fig:supp_26}
\end{figure*}

\begin{figure*}
	\centering
	\includegraphics[width=\linewidth]{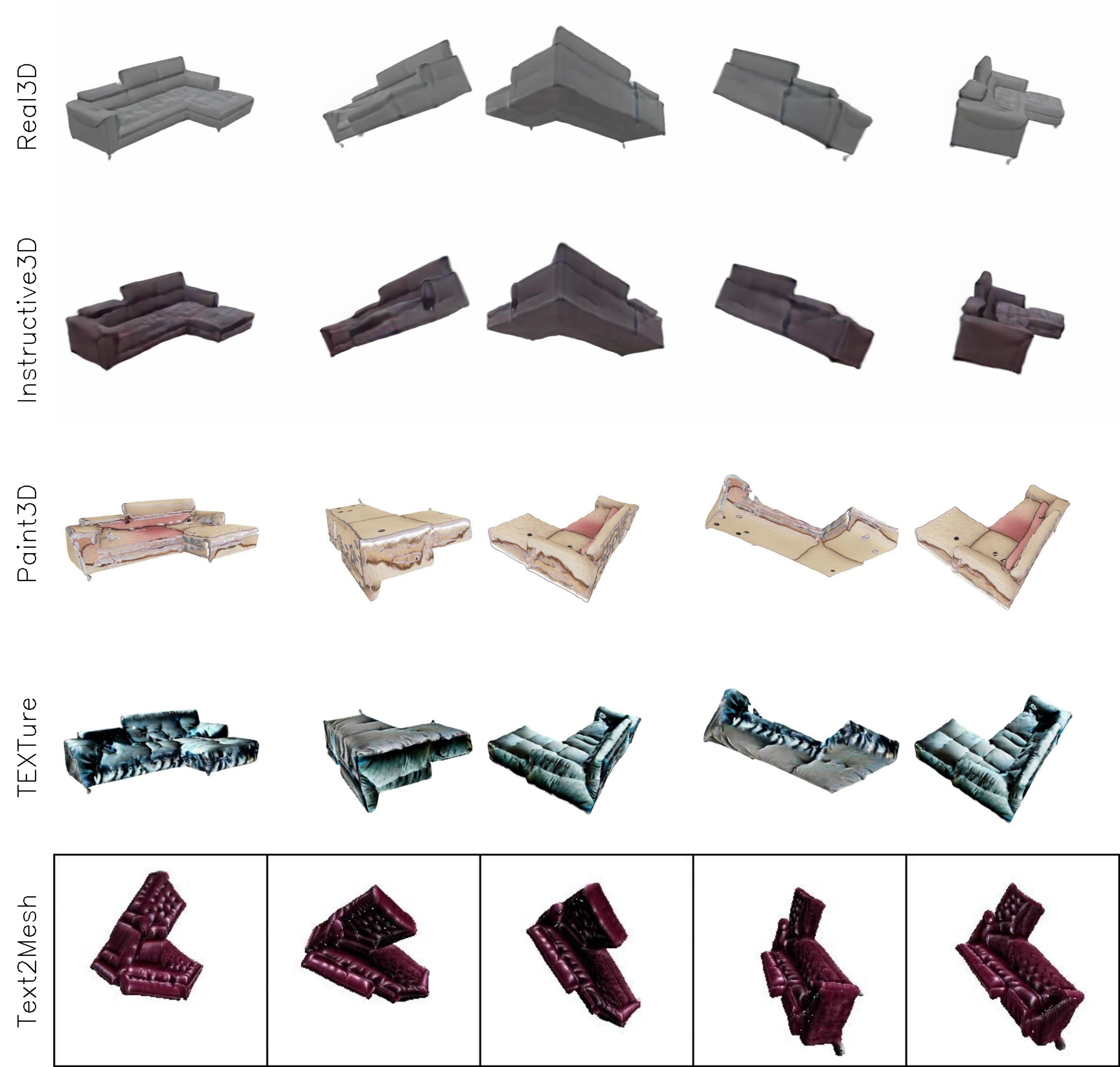}
	\caption{\textbf{Baseline comparison results.} Top row shows the rendered images from the mesh obtained from Real3D~\cite{jiang2024real3d}. Second row shows results from our method. Caption used for editing is: \textit{``darken the color of the sofa''}.}
	\label{fig:supp_27}
\end{figure*}

We compare our method with Text2Mesh~\cite{Michel2021Text2MeshTN}, Paint3D~\cite{zeng2023paint3dpaint3dlightingless} and TEXTure~\cite{richardson2023texturetextguidedtexturing3d}, which takes a 3D mesh and a text prompt as input and generates an output mesh with the given text conditioning. We provide the mesh generated by Real3D~\cite{jiang2024real3d} to these models with an edit prompt and compare the output with our generated mesh. We show additional results in Fig.~\ref{fig:supp_1}-~\ref{fig:supp_27}. These results show that our method preserves geometry and performs edits consistent with the input edit prompts.

%New figure

%New figure
%New figure

%New figure

%New figure

%New figure

%New figure

%New figure

%New figure

%New figure

%New figure

%New figure

%New figure

%New figure

%New figure

%New figure

%New figure

%New figure

%New figure

%%New figure
%\begin{figure*}
%	\centering
%	\includegraphics[width=\linewidth]{images/supp_baselines/clutch_bag_}
%	\caption{\textbf{Baseline comparison results.} Top row shows the rendered images from the mesh obtained from Real3D~\cite{jiang2024real3d}. Second row shows results from our method. Caption used for editing is: \textit{``change the color of bag to red''}.}
%	\label{fig:supp_20}
%\end{figure*}

%New figure

%New figure

%New figure

%New figure

%New figure

%New figure

%New figure

\end{document}